\documentclass[runningheads]{llncs}

\usepackage[mobile]{eccv}

\usepackage{eccvabbrv}

\usepackage{multirow}
\usepackage{graphicx}
\usepackage{booktabs}

\usepackage[accsupp]{axessibility}  

\usepackage{hyperref}

\usepackage{orcidlink}
\usepackage{soul}

\begin{document}

\title{In-the-Wild Camouflage Attack on Vehicle Detectors through Controllable Image Editing} 

\titlerunning{In-the-Wild Camouflage Attack on Vehicle Detectors}

\author{Xiao Fang\textsuperscript{1}\and 
Yiming Gong\textsuperscript{1}
    \and
    Stanislav Panev\textsuperscript{1}
    \and
    Celso de Melo\textsuperscript{2}
    \and
    Shuowen Hu\textsuperscript{2}
    \and
    Shayok Chakraborty\textsuperscript{3}
    \and
    Fernando De la Torre\textsuperscript{1}}

\authorrunning{X. Fang and Y. Gong et al.}

\institute{Carnegie Mellon University \and DEVCOM Army Research Laboratory
 \and Florida State University \\
\email{\{xfang2, yimingg2, spanev\}@andrew.cmu.edu, \{celso.m.demelo.civ, shuowen.hu.civ\}@army.mil, \ shayok@cs.fsu.edu, \ ftorre@cs.cmu.edu}
}

\maketitle

\begin{abstract}




Deep neural networks (DNNs) have achieved remarkable success in computer vision but remain highly vulnerable to adversarial attacks. Among them, camouflage attacks manipulate an object's visible appearance to deceive detectors while remaining stealthy to humans. In this paper, we propose a new framework that formulates vehicle camouflage attacks as a conditional image-editing problem. Specifically, we explore both image-level and scene-level camouflage generation strategies, and fine-tune a ControlNet to synthesize camouflaged vehicles directly on real images. We design a unified objective that jointly enforces vehicle structural fidelity, style consistency, and adversarial effectiveness. Extensive experiments on the COCO and LINZ datasets show that our method achieves significantly stronger attack effectiveness, leading to more than 38\% $\mathrm{AP}_{50}$ decrease, while better preserving vehicle structure and improving human-perceived stealthiness compared to existing approaches. Furthermore, our framework generalizes effectively to unseen black-box detectors and exhibits promising transferability to the physical world. Project page is available at \url{https://humansensinglab.github.io/CtrlCamo}


\end{abstract}    
\section{Introduction}
\label{sec:intro}


Deep neural networks (DNNs) have achieved remarkable success across a wide range of computer vision applications~\cite{resnet, segmentation, detection}. However, they are also highly vulnerable to adversarial examples that are crafted by adding carefully designed perturbations to normal examples~\cite{adversarialexamples}. For example, in the context of vehicle detection, such adversarial inputs can cause detectors to misidentify surrounding vehicles, posing serious risks to the reliability and safety of autonomous systems. As vision systems are increasingly deployed in safety-critical domains, understanding and mitigating adversarial attacks becomes crucial. 

\begin{figure}[!tb]
    \centering  
    \includegraphics[width=0.7\columnwidth]{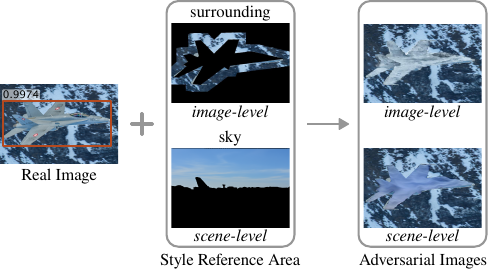}
    \caption{
    \textbf{Overview.} Given a real image, our pipeline stylizes the target vehicle based on either  its immediate surroundings (image-level) or a visual concept present in the overall scene (scene-level), producing stealthy adversarial examples. The numbers on the bounding boxes indicate detector confidence scores, and the absence of a box indicates that the vehicle is not detected.}
    \label{fig:teaser}
\end{figure}

Camouflage attack is a particular form of adversarial attack that manipulates an object’s visible appearance to deceive vision models while remaining stealthy to human observers~\cite{PhysicalAttackNaturalness}. In this work, we define ``stealthiness'' as the perceptual realism of the camouflage, referring to its ability to remain visually coherent with the scene and to avoid producing salient or unnatural patterns that attract human attention. Stealthiness is not only a desirable visual property but also a critical factor in real-world scenarios where perception and decision-making often involve human observers.  
Effective camouflage should therefore not only deceive machine detectors but also remain convincing to human observers. Moreover, evaluating attacks that preserve stealthiness provides realistic threat models, as modern detectors must be resilient not only to small pixel-level perturbations but also to plausible appearance changes that occur naturally, such as paint, wear, and decals.
In this paper, we focus on camouflage attacks that operate at the full object level to manipulate vehicle appearance to deceive vehicle detectors while maintaining stealthiness, given their wide adoption in autonomous driving, traffic management, urban planning, and defense intelligence.

Recent advances in generative AI, such as diffusion models~\cite{ddpm}, have substantially improved the fidelity and controllability of image synthesis.
These models support modular condition encoders~\cite{ControlNet} that inject structural priors such as edges and segmentation masks, enabling semantically consistent edits that respect object geometry and scene layout. Such capabilities make conditional image generation a natural fit for designing camouflage attacks that remain visually coherent with their environments while effectively deceiving object detectors.

Motivated by these observations, we approach camouflage attack as a conditional image-editing problem. As illustrated in~\cref{fig:teaser}, given a real image, our pipeline synthesizes an adversarially camouflaged image that satisfies three properties: $(i)$ preservation of the vehicle physical structure (\eg, \ the airplane in~\cref{fig:teaser}) and  surrounding background, $(ii)$ application of user-guided, stealthy style edits to the vehicle surface, and $(iii)$ reduction of detector confidence. Concretely, we fine-tune a ControlNet~\cite{ControlNet} to encode structural and stylistic guidance and optimize a composite objective that combines a structural-preservation term, a style-consistency term, and an adversarial detection loss. At inference time, our pipeline generates camouflaged images by direct sampling, and the resulting camouflage can guide camouflaging corresponding 3D real-world vehicles.  \xiao{Within this pipeline, we design two stylization strategies inspired by nature to address different practical needs. The \textit{image-level} strategy~\cite{wikipedia_crypsis} transfers visual appearance from the vehicle’s immediate surroundings, analogous to chameleons, enabling natural blending with local contexts. While effective for static imagery, this strategy is limited in real-world applications, as moving vehicles would require continual repainting across backgrounds. Therefore, we introduce a \textit{scene-level} strategy~\cite{wikipedia_mimicry}, which adapts the vehicle's appearance to a common semantic concept of the scene, analogous to grasshoppers resembling dry leaves, thereby achieving location-invariant camouflage.} For example, in~\cref{fig:teaser}, an airplane flies within a sky scene, where \textit{sky} is a common visual concept. Therefore, the pipeline adopts the blue sky as the reference style area, producing a camouflaged airplane consistent with the scene while reducing detector confidence.
Extensive experiments on both ground-view (COCO) and nadir-view (LINZ) datasets demonstrate that our pipeline achieves strong attack effectiveness, better preserves vehicle physical structure, improves stealthiness, and transfers to unseen black-box detectors and to the physical world.

Our contributions can be summarized as follows:
\begin{itemize}
    \item To the best of our knowledge, we are the first to formulate camouflage attacks against detectors on real-world images as a conditional image-editing problem and propose two camouflage strategies. The image-level strategy blends the vehicle with its  surroundings, and the scene-level strategy adapts the vehicle to a semantic concept present in the scene, producing context-aware and visually coherent camouflage.
    \item We propose a novel pipeline based on ControlNet fine-tuning. Our method jointly enforces structural fidelity to maintain vehicle geometry, style consistency to produce stealthy camouflage, and an adversarial objective to reduce detectability by object detectors.
    \item We evaluate our approach on the COCO and LINZ datasets, and demonstrate strong attack effectiveness, better preservation of vehicle physical structure, improved stealthiness, and transferability to black-box detectors and the physical world.
\end{itemize}

\section{Related Work}
\label{sec:related_work}
This section reviews prior work on camouflage  attacks. We group methods by how extensively they alter an object’s surface: imperceptible perturbations, localized patches, and full-object appearance.

\noindent\textbf{Imperceptible perturbations.} This line of work crafts small, norm-constrained perturbations applied directly to the object. Classical approaches such as TOG~\cite{TOG} add Gaussian noise and refine it iteratively to reduce detector confidence while keeping changes visually subtle. More recent techniques~\cite{diffattack, advad, advdiff, advdiffuser} employ diffusion models to inject adversarial guidance during sampling, producing minor perturbations at each step. These methods are effective for classifiers, but object detectors are generally more robust to tiny pixel-level changes, limiting the practical impact of purely imperceptible attacks on detection systems.

\noindent\textbf{Adversarial patches.} Another line of work restricts modifications to localized patches placed on the target. For example, NAP~\cite{NAP} samples patches from a pre-trained GAN and optimizes in latent space to balance stealth and attack strength. BadPatch~\cite{badpatch} uses diffusion-based inversion and mask-guided control to synthesize adversarial patches. While localized patches can achieve strong attack signals, they often introduce high-contrast patterns that contrast with the object and surroundings, making them conspicuous to human observers. 

\noindent\textbf{Full-object appearance.} The third category allows flexible modification of the entire object's appearance. A common strategy optimizes a UV texture map on a fixed 3D mesh via differentiable neural rendering, enabling gradients from adversarial objectives in image space to backpropagate to the texture~\cite{cnca, rauca, camou, uvattack}. However, this paradigm relies on precise mesh geometry, camera parameters, and lighting conditions, which are typically available only in simulation platforms~\cite{carla}. 
\xiao{Consequently, camouflages learned in simulation environments may suffer from domain gaps relative to real-world scenes, making them difficult to directly deploy on physical vehicles. Additionally, simulation environments contain a limited set of vehicle meshes and predefined scenes~\cite{synthdrive}, restricting scalability, scene diversity, and real-world transferability. In contrast, our method operates directly on in-the-wild images and generalizes flexibly across diverse scenes and vehicle types.}
Other works that operate directly on real images and combine style consistency with adversarial objectives are closer to our approach. AdvCAM~\cite{AdvCam} augments adversarial optimization with style-aware terms to align images to reference styles. DiffPGD~\cite{DiffPGD} extends this idea using diffusion priors. However, these methods target classifiers and require per-image optimization at inference. They also lack explicit constraints designed to preserve object structure or ensure scene-consistent camouflages. \xiao{In contrast, our method enforces structural fidelity, and jointly optimizes stylization and adversarial objectives to produce stealthy camouflages with efficient inference.}

\begin{figure}[!tb]
    \centering  
    \includegraphics[width=0.95\columnwidth]{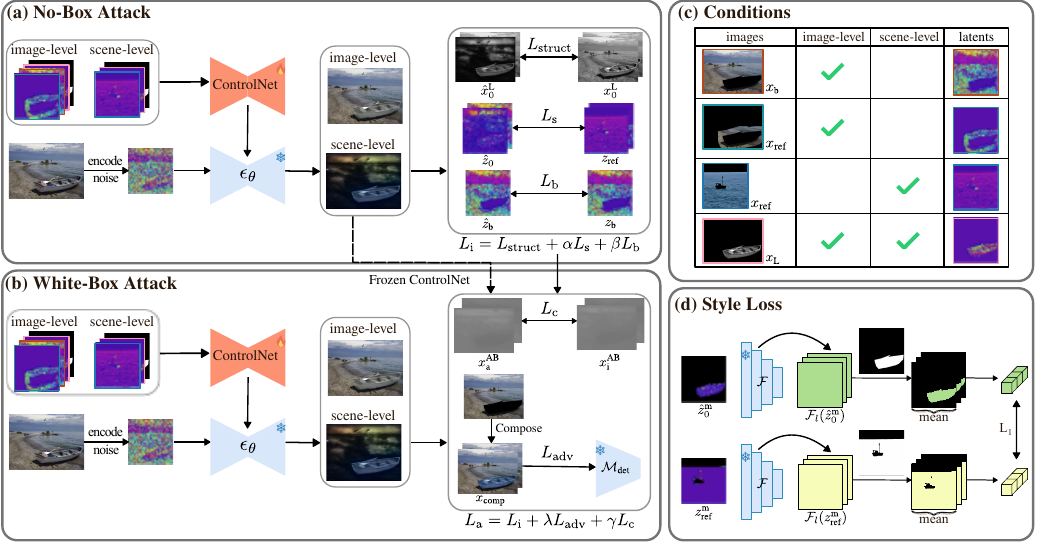}
\caption{
\textbf{Overview of our pipeline.}
As shown in (a) and (b), the pipeline consists of a \textit{No-Box Attack} stage and a \textit{White-Box Attack} stage. In (a), the ControlNet is fine-tuned to stylize vehicles using a reference region while preserving geometry and background through structure, style, and background supervisions (\(L_\text{struct}\), \(L_\text{s}\), \(L_\text{b}\)). (b) further optimizes the model against a detector \(\mathcal{M}_\text{det}\) by incorporating an additional adversarial loss \(L_\text{adv}\) and a color-consistency loss \(L_\text{c}\). (c) summarizes the conditions provided to ControlNet under the image-level and scene-level settings, and (d) illustrates the  style loss $L_\text{s}$ that aligns vehicle latent features with the reference area. 
}
    \label{fig:pipeline}
\end{figure}

\section{Method}
\label{sec:method}
We propose a two-stage framework for generating stealthy camouflage patterns that mislead vehicle detectors, while maintaining visual harmony with the surrounding environment, as shown in~\cref{fig:pipeline}. In \cref{subsec:preliminaries}, we outline the mathematical foundations of our approach. Next, \cref{subsec:stylization} introduces our image-level and scene-level stylization strategies, which automatically select appropriate style exemplars for vehicle appearance transfer. Training is performed in two sequential stages.
In the first stage (\cref{subsec:no-box attack}), termed the No-Box Attack \cite{no-box}, we fine-tune a ControlNet \cite{ControlNet} to transfer the selected reference style onto the vehicle while preserving its geometric structure, without relying on detector-dependent loss.
In the second stage (\cref{subsec:white box attack}), the model is further fine-tuned under a white-box attack setting, incorporating an adversarial objective that directly targets a known detector. This stage minimizes detectability while enforcing color and style consistency, ensuring that the adversarial camouflage retains the visual realism and stylistic attributes learned in the first stage. At inference, the trained pipeline synthesizes camouflaged images without per-image optimization.

\subsection{Preliminaries}
\label{subsec:preliminaries}
Diffusion Models~\cite{ddpm} formulate image generation as a denoising process that gradually transforms random Gaussian noise into a sample from the target data distribution. 
In this work, we employ Stable Diffusion~\cite{stablediffusion2}, which performs generation in the latent space of a pre-trained autoencoder. The model consists of an image encoder $\mathcal{E}$, a denoising network $\epsilon_\theta$, and an image decoder $\mathcal{D}$. An image $x_0$ is first encoded into the latent representation $z_0 = \mathcal{E}(x_0)$. 
The forward process  gradually adds Gaussian 
noise $\varepsilon_t$ to $z_0$:
\begin{equation}
\label{eqn:1}
    z_t =  \sqrt{\Bar{\alpha}_t} z_0 + \sqrt{1-\Bar{\alpha}_t} \varepsilon_t, \ \varepsilon_t \sim \mathcal{N}(0, \mathbf{I}), 
\end{equation} 
where $\Bar{\alpha_t}$ controls the noise schedule.
To learn the reverse process,
the network $\epsilon_\theta (z_t, c)$ is trained to predict the noise $\varepsilon$ given the noisy latent $z_t$ and a condition $c$. 
Since adversarial loss is defined on the image space, inspired by~\cite{turbofill}, we adopt a one-step estimate from the noisy latent $z_t$ to approximate the reverse process based on~\cref{eqn:1}:
\begin{equation}
\label{eqn:2}
    \hat{z}_0 = \frac{z_t - \sqrt{1-\Bar{\alpha}_t}\epsilon_\theta (z_t, c)}{\sqrt{\Bar{\alpha}_t}} \\
\end{equation}
Finally, the reconstructed image $\hat{x}_0$ is produced by decoding the estimated latent $\hat{z}_0$ through the decoder $\mathcal{D}$, which can be formulated as $\hat{x}_0 = \mathcal{D}(\hat{z}_0)$. 


\subsection{Style Reference Selection}
\label{subsec:stylization}

Our goal is to generate camouflage patterns that deceive detectors but are visually consistent with the surrounding environment. To achieve this, we introduce a process that selects a reference region serving as a style exemplar to guide the vehicle’s appearance in each image. We denote this region as 
$x_\text{ref}$ . 
In this work, we propose two complementary strategies for exemplar selection and stylization: \textit{image-level} and \textit{scene-level} camouflage generation.

In the \textit{image-level} scenario, given an input image $x_0$, the goal is to adapt the vehicle's appearance to match the style of its immediate surroundings. Let $m_{x_0}$ denote the segmentation mask of the vehicle. We first dilate the mask to obtain $m'_{x_0} = \text{dilate}(m_{x_0})$, thereby including a small region around the vehicle. The reference area is then defined as the surrounding context of the vehicle $x_\text{ref} =  x_0 \odot (m'_{x_0} - m_{x_0})$, which captures pixels adjacent to the vehicle region. 

In the \textit{scene-level} scenario, we first categorize all images into distinct scene groups using Multimodal Large Language Models (MLLMs) ~\cite{internvl3, moondream}, which infer the scene type for each image. 
For each category, we query MLLMs to identify a concept that naturally exists in the scene, ensuring that the stylized vehicle appearance remains visually consistent with real-world contexts. We then synthesize an exemplar image 
$x_\text{gen}$ containing that concept using a Stable Diffusion fine-tuned on the entire dataset, extract its segmentation mask $m_\text{gen}$ using SAM 2~\cite{sam2}, and define the reference area as $x_\text{ref} = x_\text{gen} \odot m_\text{gen}$, which captures the visual appearance of the selected representative concept. During camouflage generation, the vehicle is stylized to align with this reference, producing scene-consistent camouflage. \xiao{More implementation details and a concrete example of the entire process are provided in  Appendix~\cref{sec:supp_reference_selection}}.


\subsection{No-Box Attack}
\label{subsec:no-box attack}


In the first stage, we fine-tune a ControlNet to enable the pipeline to camouflage the vehicle using a reference style image. The pipeline takes as input the vehicle’s luminance (L) channel \(x_\text{L}\) in LAB space, the style reference region \(x_\text{ref}\) defined in~\cref{subsec:stylization}, the vehicle mask \(m_x\), and, for the image-level strategy, an additional background image \(x_\text{b}\), as shown in~\cref{fig:pipeline}(c). Given an input image \(x_0\), the estimated latent \(\hat{z}_0\) and reconstructed image \(\hat{x}_0\) are obtained from~\cref{eqn:2}. The training objective combines three components: a structure preservation loss \(L_\text{struct}\) to maintain vehicle physical structure, a style loss \(L_\text{s}\) to guide camouflage generation, and a background supervision \(L_\text{b}\) for image-level strategy. The overall loss function is formulated as
\begin{equation}
\label{eqn:3}
    L_\text{i} = L_\text{struct} + \alpha L_\text{s} + \beta L_\text{b}
\end{equation}
where $\beta = 0$ for scene-level strategy and  $\beta \neq 0$ for image-level strategy.
Next, we discuss each loss term in detail. 

\noindent\textbf{Structure preservation loss.} Inspired by colorization tasks~\cite{colorization2016} that convert the input image into LAB space and use the L channel to preserve the structure, we also utilize the vehicle's L channel for constraining the reconstructed vehicle structure. Given an input image $x_0$ and the one-step estimated image $\hat{x}_0$, we extract both L channels from ${x_0}$ and $\hat{x}_0$, and normalize them to (0,1), which we denote as $x_0^\text{L}$ and $\hat{x}_0^\text{L}$. Given the vehicle segmentation mask $m_{x_0}$, the structure preservation loss is formulated as the average $L_{2}$ difference of L channel of vehicle area between ${x_0}$ and $\hat{x}_0$:
\begin{equation}
    L_\text{struct} = \frac{1}{{\sum m_{x_0}}}\left\|x_0^\text{L} \odot m_{x_0} - \hat{x}_0^\text{L} \odot m_{x_0}\right\|^2_2.
\end{equation}

\noindent\textbf{Style loss.} The style loss is formulated based on LatentLPIPS~\cite{latentlpips}, an extension of LPIPS~\cite{lpips} that measures perceptual distance in the latent space. Prior work~\cite{gatysstyle} has shown that features learned by pre-trained classifiers effectively encode style-related information, making them well-suited for modeling perceptual style similarity. Unlike LPIPS, which compares image features in pixel space, LatentLPIPS trains a VGG~\cite{vgg} classifier 
$\mathcal{F}$ directly on diffusion latents. In this paper, we employ the pre-trained classifier $\mathcal{F}$ from LatentLPIPS instead of LPIPS because of its two advantages:  first, it is more efficient in computation and memory, as it operates in the latent space rather than pixel space; and second, it employs random differentiable augmentations such as cutout~\cite{cutout} during training, which enables more reliable comparison between masked latent regions extracted from different spatial contexts or images. This property is useful when transferring style from a reference area to a vehicle located in a different position. 

Concretely, as shown in~\cref{fig:pipeline} (d), given the one-step estimated image $\hat{x}_0$ from an input image $x_0$ via~\cref{eqn:2}, the corresponding vehicle segmentation mask $m_{x_0}$, the style image $x_\text{s}$, and its style reference area segmentation mask $ m_\text{s}$, we first encode the masked images into latent representations to suppress interference from irrelevant regions, which can be formulated as $\hat{z}_0 = \mathcal{E}(\hat{x}_0 \odot m_{x_0})$ and $z_\text{ref} = \mathcal{E}(x_\text{s} \odot  m_\text{s})$. We denote the downsampled masks as $m_{x_0\downarrow}$ and $m_{\text{s}\downarrow}$. Since zero-valued pixels do not necessarily yield zero latent activations, we further apply the downsampled masks to the latent codes to remove background interference, which can be formulated as $\hat{z}^\text{m}_0 = \hat{z}_0 \odot m_{\hat{x}_0\downarrow}$ and $z^\text{m}_\text{ref} = z_\text{ref} \odot m_{\text{s}\downarrow}$, where $m_{x_0\downarrow}$ and $m_{\text{s}\downarrow}$ have the same resolution as their corresponding latents.
Because the vehicle and style reference area occupy distinct spatial locations and may originate from different images, direct feature-wise subtraction is infeasible. Instead, for layer $l$, we extract feature maps $\mathcal{F}_l(\hat{z}^\text{m}_0)$ and $\mathcal{F}_l(z^\text{m}_\text{ref})$, and use  downsampled masks $m_{x_0\downarrow}$ and $m_{\text{s}\downarrow}$ to select the vehicle and reference regions from these feature maps. We then minimize the $L_{1}$ difference in average features within the two regions:
\begin{equation}
\label{eqn:5}
    L_\text{s} = \sum_l \left\| \frac{\sum\mathcal{F}_l(\hat{z}^\text{m}_0)[m_{x_0\downarrow}]}{\sum m_{x_0\downarrow}} - \frac{\mathcal{F}_l(z^\text{m}_\text{ref}) [m_{\text{s}\downarrow}]}{\sum m_{\text{s}\downarrow}} \right\|
\end{equation}
where $m_{x_0\downarrow}$ and $m_{s\downarrow}$ are resized to match the spatial resolution of the feature map at each layer $l$.

\noindent\textbf{Background reconstruction loss.} We observe that reconstructing the background leads to more coherent vehicle stylization under the image-level strategy, where the vehicle appearance is expected to be aligned with its immediate surroundings. This is because in the image-level setting, each image is conditioned on its own surroundings rather than a few shared reference areas, which makes style transfer more challenging via average feature-space loss. Background supervision introduces stronger pixel-level constraints that anchor the global image color and illumination distribution, allowing gradients to propagate through shared features and harmonize the vehicle's appearance with its surroundings. Similarly, given the one-step estimated image $\hat{x}_0$ from an input image $x_0$ via~\cref{eqn:2}, the corresponding vehicle segmentation mask $m_{x_0}$, we encode the masked images into latent representations to suppress interference from vehicle regions, which can be formulated as $z_\text{b} = \mathcal{E}(x_0 \odot (1-m_{x_0}))$ and $\hat{z}_\text{b} = \mathcal{E}(\hat{x}_0 \odot (1-m_{x_0}))$. We further apply the downsampled masks $m_{x_0\downarrow}$ to $z_\text{b}$ and $\hat{z}_\text{b}$ to focus on the background, which can be formulated as $z^\text{m}_\text{b} = z_\text{b} \odot (1-m_{x_0\downarrow})$ and $\hat{z}^\text{m}_\text{b} = \hat{z}_\text{b} \odot (1-m_{x_0\downarrow})$.  Background reconstruction loss $L_\text{b}$ is formulated to be the LatentLPIPS~\cite{latentlpips} loss between $z^\text{m}_\text{b}$ and $\hat{z}^\text{m}_\text{b}$, which minimizes both latent-space pixel-wise and perceptual feature differences. 

\subsection{White-Box Attack}
\label{subsec:white box attack}
In the second stage, we continue to fine-tune the ControlNet from the first stage as described in~\cref{subsec:no-box attack}. The goal is to preserve the vehicle appearance learned in the first stage while deceiving the vehicle detector $\mathcal{M}_\text{det}$. Concretely, we augment the first-stage loss \(L_{\mathrm{i}}\) from~\cref{eqn:3} with two terms: a color-consistency loss \(L_{\mathrm{c}}\) that constrains chromatic deviation in the adversarial output, and an adversarial detection loss \(L_{\mathrm{adv}}\). The combined objective is formulated as
\begin{equation}
\label{eqn:6}
    L_\text{a} = L_\text{i} + \lambda L_\text{adv} + \gamma L_\text{c}
\end{equation}
Next, we discuss each loss term in detail.

\noindent\textbf{Adversarial loss.} Given the one-step estimated image $\hat{x}_0$ from an input image $x_0$ via~\cref{eqn:2}, and vehicle segmentation mask $m_x$, since for camouflage attack we are only allowed to edit the vehicle, we compose real image background and estimated image vehicle before passing it through a detector, which can be formulated as $x_\text{comp} = \hat{x}_0 \odot m_x + x_0 \odot (1-m_x)$. We then optimize the camouflaged vehicle to be detected as background by the detector. Formally, if $\mathcal{M}_\text{det}(x_\text{comp})$ denotes the detector logits, and $y_\text{b}$ is the background label, the adversarial objective can be written as a cross-entropy loss:
\begin{equation}
    L_\text{adv} = \text{CE} (\mathcal{M}_\text{det}(x_\text{comp}), y_\text{b})
    \label{eq:adv_loss}
\end{equation}

\noindent\textbf{Color-consistency loss.} While the style loss $L_\text{s}$ introduced in~\cref{subsec:no-box attack} aligns the feature representation between the vehicles in generated images and the reference area, we observe that during white-box attacks, the model may slightly shift vehicle colors while keeping the style loss nearly unchanged to facilitate optimization of the adversarial loss $L_\text{adv}$, leading to undesired color deviations. To address this issue, inspired by DINOv3~\cite{dinov3}, we introduce a color-consistency loss that leverages the knowledge from the previous training stage to stabilize vehicle appearance. Specifically, we condition on the frozen ControlNet trained in~\cref{subsec:no-box attack} and the ControlNet trained in this stage to reconstruct one-step outputs from~\cref{eqn:2}, denoted as $x_\text{i}$ and $x_\text{a}$, respectively. Both outputs are converted to the LAB color space, and we extract normalized AB channels, yielding $x_\text{i}^\text{AB}$ and $x_\text{a}^\text{AB}$, and their difference is minimized to ensure consistent color representation across stages. Given the vehicle segmentation mask $m_{x_0}$, the loss is computed as the mean $L_{2}$ distance in the AB channels over vehicle regions:
\begin{equation}
    L_\text{c} = \frac{1}{{\sum m_{x_0}}}\left\|x_\text{i}^\text{AB} \odot m_{x_0} - x_\text{a}^\text{AB} \odot m_{x_0}\right\|^2_2
\end{equation}
\section{Experiments}
\label{sec:experiments}

We conduct comprehensive experiments to evaluate the effectiveness of our method from four perspectives: attack effectiveness, style stealthiness, structural preservation, and transferability. In~\cref{subsec:setup}, we describe the setup. In~\cref{subsec:sota}, we compare our approach with state-of-the-art methods under the white-box setting, where target detectors are known. In~\cref{subsec:transferability}, we assess the transferability of our method to both black-box settings, where the victim models are unknown, and the physical world. In~\cref{subsec:ablation}, we conduct ablation studies.

\subsection{Experimental Setup}
\label{subsec:setup}

\noindent\textbf{Datasets.}
We conducted experiments on two public datasets: LINZ~\cite{LINZ} and COCO~\cite{coco}.
The LINZ dataset consists of aerial imagery collected over Selwyn, New Zealand, with a ground sampling distance (GSD) of 12.5 cm per pixel. Each image is cropped to a resolution of $112 \times 112$ pixels and annotated with pseudo–bounding boxes marking car centers. After filtering images that contain cars, the dataset includes $6,011$ training and $728$ testing samples. The scenes are categorized into five types, namely \textit{residential, industrial, agricultural, parking lot,} and \textit{highway}. Each category is associated with a corresponding style reference concept (\textit{house, building, field, tree,} and \textit{grass}) used for scene-level stylization. The COCO dataset contains diverse natural scenes with complex object interactions. From this dataset, we extract images containing a vehicle, resulting in $8,965$ training samples and $400$ testing samples. The data are grouped into five environments: \textit{urban, rural, road, sky}, and \textit{lake}. Each environment is paired with a representative style reference concept (\textit{building, grass, tree, sky}, and \textit{water}) that guides the stylization process. \xiao{Following previous texture-based attack methods~\cite{rauca, cnca, camou, uvattack}, we assume access to vehicle regions, where ground-truth vehicle masks
are provided in both datasets.}

\noindent\textbf{Implementation.} We adopt Stable Diffusion v1.5~\cite{stablediffusion2} as our generative model and fine-tune ControlNet~\cite{ControlNet} in both stages with a batch size of $4$ on two RTX A6000 GPUs. 
The dilation kernel size is set to 75~px for COCO and 11~px for LINZ. 
The images are all resized to 512~px $\times$ 512~px. We use a template prompt of the form ``an image of \{scene type\} area with \{objects\}'', where \{scene type\} is the scene label of the image and \{objects\} lists the objects present in the image. At test time, we run the pipeline for $30$ sampling steps. For attacks and evaluation, we use the MMDetection framework~\cite{mmdetection} with Faster-RCNN~\cite{faster-rcnn} and ViTDet~\cite{vitdet} as the
white-box target detection models for adversarial camouflage generation. To evaluate the transferability of our method, we further treat YOLOv5~\cite{yolov5}, YOLOv8~\cite{yolov8}, and MLLMs~\cite{moondream, internvl3} as black-box models and test on them. Attack effectiveness is reported as the drop in $\mathrm{AP}_{50}$ relative to baseline detectors. To assess the preservation of vehicle structure, we report the Structural Similarity Index Metric (SSIM). Input images are cropped to reduce the background and enlarge the vehicle. Finally, we measure inference latency (seconds per image) to characterize average adversarial image generation time across the dataset. \xiao{More hyperparameter details are discussed in the Appendix~\cref{subsec:supp_setup}.}

\begin{table}[!tb]
\caption{\textbf{Quantitative comparison on the LINZ Dataset for stylization-based camouflage attacks}. We report the $\mathrm{AP}_{50}$, SSIM, and average sampling time.}

\resizebox{\textwidth}{!}{
\begin{tabular}{c|ccccc|ccccc}
\hline
\multirow{3}{*}{Method} & \multicolumn{5}{c|}{Faster-RCNN}   & \multicolumn{5}{c}{ViTDet}     \\ \cline{2-11} 
& \multicolumn{2}{c}{image-level} & \multicolumn{2}{c|}{scene-level}      & \multirow{2}{*}{Inf. Latency (s) $\downarrow$} & \multicolumn{2}{c}{image-level} & \multicolumn{2}{c|}{scene-level}   & \multirow{2}{*}{Inf. Latency (s) $\downarrow$} \\ \cline{2-5} \cline{7-10}
& $\mathrm{AP}_{50} (\%) \downarrow$ & SSIM $\uparrow$ & $\mathrm{AP}_{50} (\%) \downarrow$ & \multicolumn{1}{c|}{SSIM $\uparrow$} &        & $\mathrm{AP}_{50} (\%) \downarrow$   & SSIM $\uparrow$  & $\mathrm{AP}_{50} (\%) \downarrow$ & \multicolumn{1}{c|}{SSIM $\uparrow$} &     \\ \hline
Normal &   98.3   &     - &   98.3        &   \multicolumn{1}{c|}{-}   &   -        &  97.8       &     -       &  97.8      & \multicolumn{1}{c|}{-}   &  -   \\
AdvCAM~\cite{AdvCam}   &    90.5     &    0.968   &    88.6       &   \multicolumn{1}{c|}{\textbf{0.963}}                        &  21.8  & 84.1                        &   0.964               &       80.8     & \multicolumn{1}{c|}{0.963}     &    21.8  \\
Diff-PGD~\cite{DiffPGD}   &    73.2   &       0.966 &     66.4      &    \multicolumn{1}{c|}{0.959}        &  32.0 & 62.3   & 0.961  &  59.6  &       \multicolumn{1}{c|}{0.96}     &  32.4   \\
Ours   &   \textbf{18.3}  & \textbf{0.972}  &   \textbf{27.5}   &  \multicolumn{1}{c|}{0.961}     &    \textbf{7.00}    &     \textbf{13.7}   &    \textbf{0.972}     &    \textbf{11.1}   & \multicolumn{1}{c|}{\textbf{0.964}}     & \textbf{7.67}   \\ \hline
\end{tabular}}

\label{tab:sota-linz}
\end{table}

\begin{table}[!tb]

\caption{\textbf{Quantative comparison on the COCO Dataset for stylization-based camouflage attacks}. We report the $\mathrm{AP}_{50}$, SSIM, and average sampling time.}

\resizebox{\textwidth}{!}{
\begin{tabular}{c|ccccc|ccccc}
\hline
\multirow{3}{*}{Method} & \multicolumn{5}{c|}{Faster-RCNN}   & \multicolumn{5}{c}{ViTDet}     \\ \cline{2-11} 
& \multicolumn{2}{c}{image-level} & \multicolumn{2}{c|}{scene-level}      & \multirow{2}{*}{Inf. Latency (s) $\downarrow$} & \multicolumn{2}{c}{image-level} & \multicolumn{2}{c|}{scene-level}   & \multirow{2}{*}{Inf. Latency (s) $\downarrow$} \\ \cline{2-5} \cline{7-10}
& $\mathrm{AP}_{50} (\%) \downarrow$ & SSIM $\uparrow$ & $\mathrm{AP}_{50} (\%) \downarrow$ & \multicolumn{1}{c|}{SSIM $\uparrow$} &        & $\mathrm{AP}_{50} (\%) \downarrow$   & SSIM $\uparrow$  & $\mathrm{AP}_{50} (\%) \downarrow$ & \multicolumn{1}{c|}{SSIM $\uparrow$} &     \\ \hline
Normal &   85.6   &     - &   85.6       &   \multicolumn{1}{c|}{-}   &   -        &  91.4       &     -      &  91.4      & \multicolumn{1}{c|}{-}   & -  \\
AdvCAM~\cite{AdvCam}   &    64.7     &    0.680   &    66.4       &   \multicolumn{1}{c|}{0.651}                        &   21.8 &  70.5                        &  0.678               &       68.3    & \multicolumn{1}{c|}{0.650}     &   21.8    \\
Diff-PGD~\cite{DiffPGD}   &    61.1   &       0.692 &     64.6      &    \multicolumn{1}{c|}{0.664}        & 33.2 & 68.5  & 0.677  &  68.5  &       \multicolumn{1}{c|}{0.663}     &    34.2 \\
Ours   &   \textbf{15.0}  & \textbf{0.850}  &   \textbf{16.6}   &  \multicolumn{1}{c|}{\textbf{0.837}}     &    \textbf{7.32}    &     \textbf{19.2}   &    \textbf{0.849}     &    \textbf{12.5}   & \multicolumn{1}{c|}{\textbf{0.840}}     &  \textbf{8.15}  \\ \hline
\end{tabular}}   
\label{tab:sota-coco}
\end{table}

\setlength{\tabcolsep}{2pt}
\begin{table}[!tb]
\caption{\textbf{Quantitative comparison with non-stylization camouflage attacks.} ``Noise'' and ``Patch'' denote adding random perturbations and patches on vehicles to conduct camouflage attack.  We report the average $\mathrm{AP}_{50}$ of both strategies.}

\centering
\footnotesize
\begin{tabular}{c|c|cc|cccc}
\hline
\multirow{3}{*}{Method} & \multirow{3}{*}{Type} & \multicolumn{2}{c|}{LINZ} & \multicolumn{2}{c}{COCO}  \\ \cline{3-6} &  & Faster-RCNN  & ViTDet     & Faster-RCNN & ViTDet \\ \cline{3-6} &                       & $\mathrm{AP}_{50}  (\%) $  & $\mathrm{AP}_{50} (\%) $       & $\mathrm{AP}_{50} (\%) $        & $\mathrm{AP}_{50} (\%) $ \\ \hline
Normal  & -     &   98.3           &     97.8       &   85.6             &  91.4  \\
ToG~\cite{TOG}  & Noise & 83.5     &   82.8         &   72.5            &       80.0                       \\
DiffAttack~\cite{diffattack}   & Noise & 98.0      &   97.5         &       77.7        &            83.2                  \\
NAP~\cite{NAP} & Patch                 &      89.2        &       88.6     &       75.4        &   87.1                     \\
BadPatch~\cite{badpatch}  & Patch & 96.5      &       88.7     &        73.8       &                84.4              \\
Ours      &  Style                & \textbf{22.9}      &     \textbf{12.4}       &     \textbf{15.8}       &      \textbf{15.9}   \\ \hline
\end{tabular}
\label{tab:sota-other-types}
\end{table}

\begin{figure}[!tb]
    \centering  
    \includegraphics[width=0.98\columnwidth]{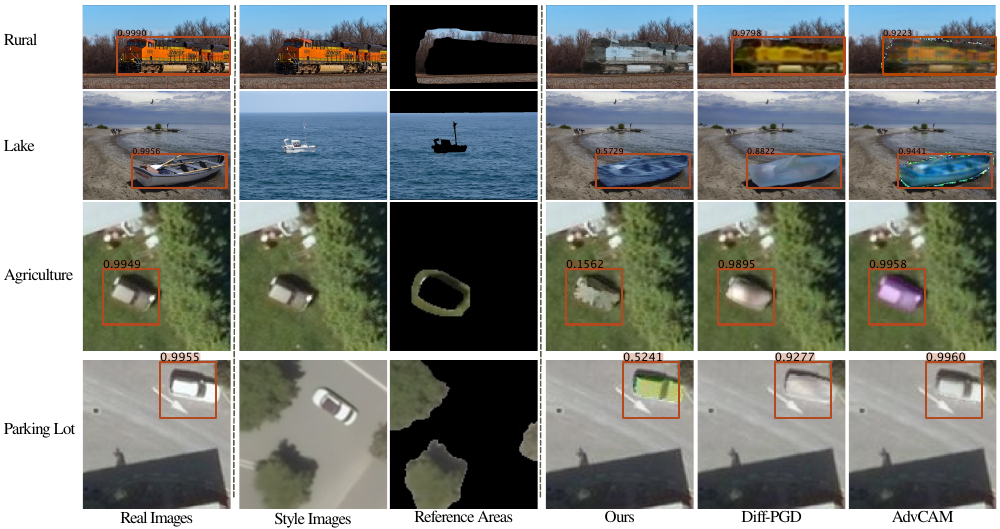}
    \caption{\textbf{Qualitative comparison with other methods.} The first two rows show results from the COCO dataset, and the last two rows are from the LINZ dataset. Within each dataset, the first row corresponds to the image-level strategy, and the second row corresponds to the scene-level strategy. Scene types are indicated on the left. In the ``lake'' scene,  boats are stylized toward the water, while in the ``parking lot'' scene, cars are stylized toward trees.  All camouflaged images are composed with real images background.}
    \label{fig:sota}
\end{figure}

\subsection{Comparison with State-of-the-art Methods}
\label{subsec:sota}

We compare our approach with two categories of state-of-the-art methods. The first category comprises stylization-based approaches that, similar to ours, camouflage the vehicle’s appearance to reference areas. As shown in~\cref{tab:sota-linz} and~\cref{tab:sota-coco}, our method achieves at least $38.9\%$ $\mathrm{AP}_{50}$ reduction compared to both AdvCAM~\cite{AdvCam} and Diff-PGD~\cite{DiffPGD} across datasets,  detectors, and strategies, demonstrating a consistently stronger attack performance.  In addition, our method achieves a significantly higher SSIM score on COCO, demonstrating superior preservation of vehicle structure. Moreover, these methods require per-image optimization during camouflage generation, leading to much higher inference times than our pipeline.

\Cref{fig:sota} presents qualitative comparisons evaluated on Faster-RCNN across diverse environments, including rural, sky, agricultural, and parking lot scenes. Our approach effectively stylizes the vehicles based on the visual characteristics of the reference area while preserving their geometric structure. In contrast, Diff-PGD and AdvCAM rely on square patches, resulting in less coherent stylization and weaker correspondence to scene context. By introducing a spatially adaptive stylization process, our method achieves faithful style transfer to the reference area and produces visually consistent camouflage.

\xiao{We further assess stealthiness through a human study. In our formulation, stealthiness is operationalized as perceptual alignment between the camouflaged vehicle and its surrounding context or reference objects, following the stylization principles inspired by nature (see~\cref{sec:intro}). Accordingly, participants were asked to select the method whose stylization best matched the reference areas. At the image level, our method was preferred in 53.1\% of cases, compared to 36.5\% for Diff-PGD and 10.4\% for AdvCAM. At the scene level, preference for our method increased to 85.3\%, versus 11.7\% and 3.0\%, respectively, indicating improved human-perceived stealthiness under both strategies. Additional details and example questions are provided in Appendix~\cref{subsec:supp_human_study_compare_sota}.} 

The second category includes non-stylization camouflage attacks, such as adding random noise and patches, as shown in~\cref{tab:sota-other-types}. Compared to these paradigms, our pipeline achieves over $56\%$ $\mathrm{AP}_{50}$ deduction across datasets and detectors, validating the  effectiveness of our approach. 

\begin{table}[!tb]
\caption{\textbf{Evaluation of camouflages across
various models.} We report the average of image-level and scene-level strategies using $\mathrm{AP}_{50}$ for detectors, and classification accuracy for MLLMs.}

\centering
\begin{tabular}{cc|ccccc}
\hline
Dataset & Surrogate   & Faster-RCNN & ViTDet & YOLOv5 & YOLOv8 & MLLM \\ 
\hline
LINZ    & Normal & 98.3           &     97.8   & 97.4 & 95.4      &  96.3    \\
LINZ    & Faster-RCNN & -           &     33.7   & 32.3 & 19.3      &   10.7   \\
LINZ    & ViTDet      &  11.4           & -    & 38.6  &  25.2      &    11.3  \\
\hline
COCO    & Normal &  85.6           &   91.4  &  88.3 &  90.7      &   87.5   \\
COCO    & Faster-RCNN & -           &    40.0     &  30.2  &  39.0     & 31.8     \\
COCO   & ViTDet      &       38.5      & -      & 34.4 &  39.6     &  33.0    \\ 
\hline
\end{tabular}
\label{tab:transferability}
\end{table} 

\subsection{Transferability}
\label{subsec:transferability}

\noindent\textbf{Transferability to Black-Box Models.}
We evaluate black-box transferability by generating camouflages using Faster-RCNN and ViTDet to unseen targets, as shown in~\cref{tab:transferability}. Across both LINZ and COCO datasets, our method leads to at least a $47.1\%$ drop in detection $\mathrm{AP}_{50}$ and a $54.5\%$ decrease in classification accuracy when identifying vehicle presence in the image. These results indicate that the camouflage patterns are not overfitted to specific detectors but generalize across architectures, demonstrating robustness and cross-model transferability. 

\begin{figure}[!tb]
    \centering  
    \includegraphics[width=0.8\columnwidth]{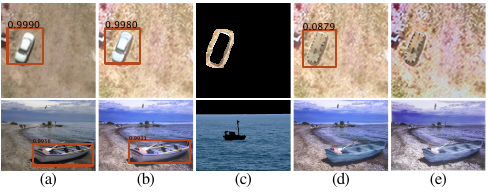}
    \caption{
    \xiao{\textbf{Projector-based physical experiment.} (a) Real images in digital space. (b) Photos captured from real images. (c) Reference areas used for style guidance.  (d) Camouflaged images generated from the captured photos.  (e) Photos taken after projecting the camouflaged images back onto the 3D physical models.
    }}
    \label{fig:projector}
\end{figure}

\noindent\textbf{Transferability to the Physical World.}
\xiao{Following~\cite{projattacker}, we conduct a projector-based experiment targeting Faster-RCNN on COCO and LINZ datasets to evaluate the real-world applicability. For COCO, we reconstruct the scenes by monocular depth estimation~\cite{depthanything3} and 3D-print them. For LINZ, we 3D-print car models resembling those in the dataset. We then project images onto the physical models attached to a white board to mimic physical surface painting and capture photos using a smartphone, as shown in~\cref{fig:projector}.}
Although this setup introduces minor illumination and color variations, vehicle detection confidence remains nearly unchanged between (a) and (b), indicating robustness to moderate appearance shifts. We then apply our pipeline to generate camouflaged images, composite them with the real-world backgrounds, as shown in~\cref{fig:projector}(d), and project the results back onto the physical models for a second photo, as shown in~\cref{fig:projector}(e). Compared with non-adversarial images, the camouflaged cases exhibit a clear drop in detection confidence, indicating that  adversarial patterns learned in simulation can effectively transfer to physical environments. \xiao{More setup details and results are reported in \cref{subsec:supp_physical_experiment}.}

\begin{figure}[!tb]
    \centering  
    \includegraphics[width=0.65\columnwidth]{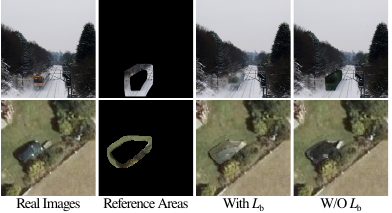}

\caption{\textbf{Effectiveness of background supervision.}}
\label{fig:background}
\end{figure}
\begin{figure}[!tb]
    \centering  
    \includegraphics[width=0.8\columnwidth]{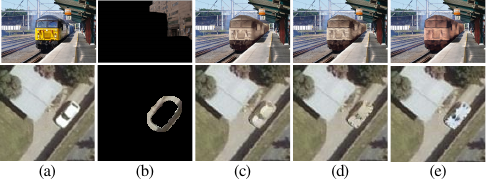}
    \caption{\textbf{Effectiveness of color-consistency loss.} (a) Real images. (b) Reference areas used for style guidance. (c) Camouflaged images generated in the ``No-Box Attack'' stage.  Camouflaged images with $L_\text{c}$. (d) Camouflaged images without $L_\text{c}$. 
    }
    \label{fig:color}
\end{figure}

\subsection{Ablation Study}
\label{subsec:ablation}


\noindent\textbf{Effectiveness of background supervision.} We assess the role of background reconstruction loss $L_\text{b}$ in the image-level setting introduced in~\cref{subsec:no-box attack}, as illustrated in~\cref{fig:background}. As observed across both datasets, incorporating $L_\text{b}$ encourages the model to stylize vehicles according to the visual characteristics of their surrounding context, resulting in more coherent and spatially consistent camouflage.

\noindent\textbf{Effectiveness of color-consistency loss.} We evaluate the effectiveness of the color-consistency loss $L_\text{c}$ introduced in~\cref{subsec:white box attack}. Incorporating $L_\text{c}$ effectively constrains the color of the generated camouflage to remain consistent with the appearance established in the first stage. As shown in~\cref{fig:color}(d) and (e) compared to (c), this loss mitigates large color deviations arising during adversarial optimization, thereby enhancing visual coherence between the stylized vehicle and its reference appearance. \xiao{Ablation studies on the two-stage paradigm and other losses are discussed in Appendix~\cref{subsec:supp_ablation}.}
\section{Conclusions}
\label{sec:conclusions}
In this paper, we formulate camouflage attack as a conditional image editing problem that produces stealthy vehicle camouflages that reduce detector performance while remaining stealthy to human observers. We introduce two complementary stylization strategies: an image-level strategy that adapts vehicle appearance to its surroundings, and a scene-level strategy that leverages visual concepts in the scene. Building on these strategies, we develop a two stage framework that jointly stylizes vehicles while enforcing structural fidelity and adversarial effectiveness. Experiments on the LINZ and COCO datasets show that our approach generates stealthier camouflages, better preserves vehicle structures, and transfers to unseen black-box detectors and the physical world.

\xiao{We also identify several limitations that point to future research directions. First, our pipeline formulates camouflage attacks as a digital image-editing problem and therefore does not explicitly model 3D geometry, material properties, or viewpoint variation, which are advantages of 3D texture-based approaches. To approximate the effect of physically camouflaging vehicles, we use the L channel in LAB color space as a coarse proxy for shading during optimization. However, in single-view images the L channel entangles illumination, geometry, and material effects, preventing reliable disentanglement of surface reflectance. As a result, our method cannot explicitly manipulate the albedo corresponding to the unshaded surface texture. In future work, we plan to explore geometry-aware representations and multi-view real-vehicle capture to enable more accurate modeling of shading and material properties.
Second, the image-level strategy is less effective for ground-view images. Due to perspective projection, the dilated surroundings of a vehicle mask often include regions that are spatially distant from the target object and may contain multiple semantically diverse categories. This makes it difficult to learn a consistent style reference from the surrounding context, leading to less coherent camouflage patterns.} 

\section*{Acknowledgements}
This work has been
funded by the DEVCOM Army Research Laboratory.

%
%

\bibliographystyle{splncs04}
\bibliography{main}

\clearpage

\setcounter{page}{1}

\title{In-the-Wild Camouflage Attack on Vehicle Detectors through Controllable Image Editing Supplementary Material} 

\titlerunning{Supplementary Material}

\author{Xiao Fang\textsuperscript{1}\and 
Yiming Gong\textsuperscript{1}
    \and
    Stanislav Panev\textsuperscript{1}
    \and
    Celso de Melo\textsuperscript{2}
    \and
    Shuowen Hu\textsuperscript{2}
    \and
    Shayok Chakraborty\textsuperscript{3}
    \and
    Fernando De la Torre\textsuperscript{1}}

\authorrunning{X. Fang and Y. Gong et al.}

\institute{Carnegie Mellon University \and DEVCOM Army Research Laboratory
 \and Florida State University \\
\email{\{xfang2, yimingg2, spanev\}@andrew.cmu.edu, \{celso.m.demelo.civ, shuowen.hu.civ\}@army.mil, \ shayok@cs.fsu.edu, \ ftorre@cs.cmu.edu}
}

\maketitle

\renewcommand{\thesection}{\Alph{section}}
\renewcommand{\theHsection}{\Alph{section}}
\renewcommand{\theHfigure}{S\arabic{figure}}
\renewcommand{\thefigure}{S\arabic{figure}}
\renewcommand{\thetable}{S\arabic{table}}
\renewcommand{\theHtable}{S\arabic{table}}

\xiao{\paragraph{Organization of the Supplementary Material.}
\begin{itemize}
    \item ~\cref{sec:supp_reference_selection} describes the procedure for selecting reference areas used in scene-level stylization strategy.
    \item ~\cref{sec:supp_rectflow} presents an extension of our framework to rectified flow.
    \item ~\cref{sec:supp_experiments} presents additional implementation details and experimental results, including qualitative examples, robustness under preprocessing defenses, and ablation studies.
    \item ~\cref{sec:supp_human} reports details of the human evaluation protocol and results presented in~\cref{subsec:sota}, and an additional human study comparing with ablation variants.
    \item ~\cref{sec:supp_transferability} presents an additional experiment on cross-location transferability, and more visualizations and quantitative results on projector-based physical tests presented in~\cref{subsec:transferability}.
    \item ~\cref{sec:supp_limitations} discusses the limitations of our method in more details.
\end{itemize}
}

\section{Style Reference Selection}
\label{sec:supp_reference_selection}
\begin{figure}[!tb]
\centering  
\includegraphics[width=0.98\textwidth]{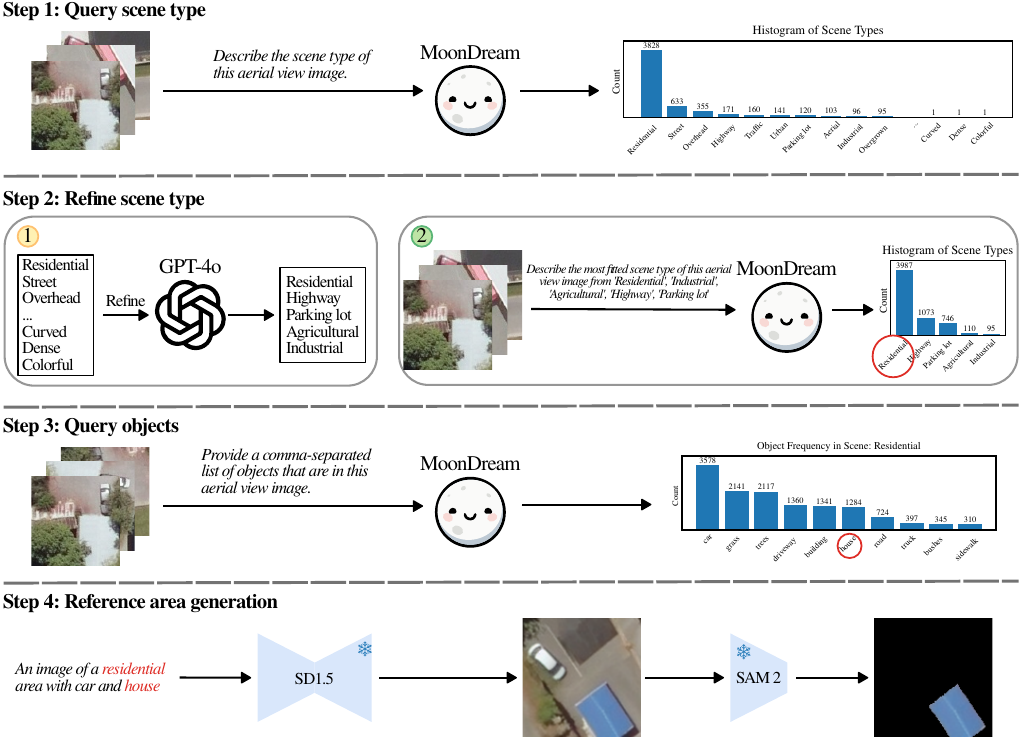}
\caption{
\textbf{Workflow for selecting style reference areas for the scene-level camouflage generation strategy.}
The process consists of four steps. In the first step, we query MoonDream for initial scene types. In the second step, we refine these categories using GPT-4o and recollect labels under five representative scene groups. In the third step, we query object-level information in each image. In the last step, we select a concept in each scene and synthesize an exemplar image containing that concept. We then segment the target concept with SAM~2. The extracted region serves as the reference area for scene-level stylization. Red circles denote we use the ``house'' concept in ``Residential'' area as an example to illustrate how we generate reference area in the last step.
}
\label{fig:supp_scene_level_process}
\end{figure}

We describe our procedure for selecting reference areas for the \textit{scene-level} stylization strategy (see~\cref{subsec:stylization}) using an example workflow on the LINZ dataset. As illustrated in~\cref{fig:supp_scene_level_process}, the process consists of four stages.

In the first step, we query MoonDream~\cite{moondream} with each image in LINZ using the prompt ``\textit{Describe the scene type of this aerial view image.}'' to obtain an initial scene-type prediction. These raw responses, however, are noisy and contain numerous fine-grained or synonymous categories.

To consolidate these labels, the second step refines the scene types. We feed the full list of MoonDream-generated categories into GPT-4o~\cite{gpt4o} with the prompt ``\textit{Select a subset of scene groups that cover a wide variety of scene types and minimize semantic overlap between each category.}'' GPT-4o clusters the categories into a compact set of five representative classes: \textit{Residential, Industrial, Agricultural, Highway, and Parking lot} (Fig.~\ref{fig:supp_scene_level_process}, Step 2(1)). We then re-query MoonDream with the constrained prompt ``\textit{Describe the most fitted scene type of this aerial view image from [Residential, Industrial, Agricultural, Highway, Parking lot].}'' to assign each image to one of these refined categories (Fig.~\ref{fig:supp_scene_level_process}, Step 2(2)).

After determining the scene type for every image, the third step extracts object-level information. We again query MoonDream with the prompt ``\textit{Provide a comma-separated list of objects that are in this aerial view image.}'' and aggregate the distributions within each scene group. An example object-frequency histogram for the \textit{Residential} category is presented in~\cref{fig:supp_scene_level_process}. 

Finally, in the last step, we identify a representative concept for each refined scene type and synthesize a reference exemplar image. We fine-tune Stable Diffusion (SD) v1.5 on the LINZ dataset following the template prompt ``\textit{an image of {scene type} area with {objects}}'' described in~\cref{subsec:setup}. Within each scene group, we select a common concept and generate an image containing that concept. For instance, as shown in~\cref{fig:supp_scene_level_process}, for \textit{Residential} scene, we select the concept \textit{house} and synthesize an image using the prompt ``\textit{An image of a residential area with car and house}.'' We then apply SAM~2~\cite{sam2} to segment and extract the spatial region corresponding to the chosen concept, which serves as the reference area for \textit{scene-level} stylization.

\begin{figure}[!tb]
    \centering  
    \includegraphics[width=0.75\columnwidth]{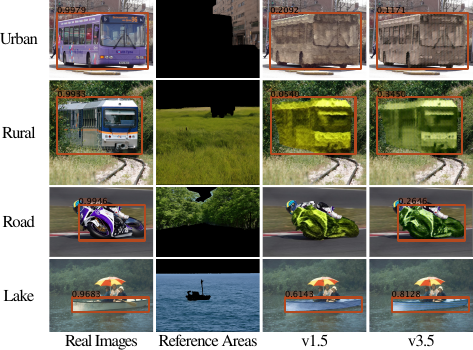}
    \caption{\textbf{Qualitative evaluation of SD v3.5 on COCO.} We visualize scene-level camouflage generation results. Each row corresponds to a scene type. Numbers above the bounding boxes indicate detector confidence, and the absence of a box indicates a missed detection. All camouflaged vehicles are composited onto the original real-image backgrounds.
    }
    \label{fig:supp_scene_level_sd35}
\end{figure}
\begin{figure}[!tb]
    \centering  
    \includegraphics[width=0.75\columnwidth]{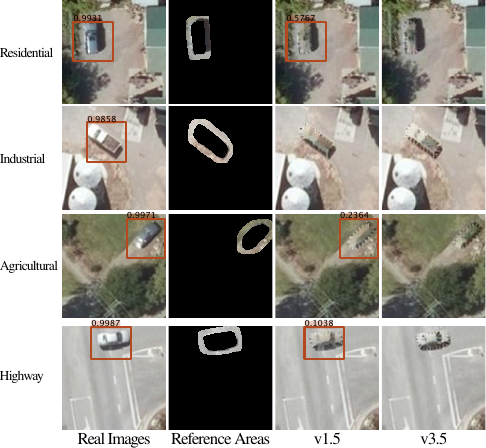}
    \caption{\textbf{Qualitative evaluation of SD v3.5 on LINZ.} We visualize image-level camouflage generation results. 
    }
    \label{fig:supp_image_level_sd35}
\end{figure}

\begin{table}[!tb]
\caption{\textbf{Quantative comparison between SD v1.5 and v3.5.} The \textit{image-level} strategy is evaluated on the LINZ dataset, while the \textit{scene-level} strategy is evaluated on the COCO dataset. We report the $\mathrm{AP}_{50}$, SSIM, and average sampling time per image.}

\resizebox{\textwidth}{!}{
\begin{tabular}{c|ccccc|ccccc}
\hline
\multirow{3}{*}{Method} & \multicolumn{5}{c|}{Faster-RCNN}   & \multicolumn{5}{c}{ViTDet}     \\ \cline{2-11} 
& \multicolumn{2}{c}{image-level} & \multicolumn{2}{c|}{scene-level}      & \multirow{2}{*}{Inf. Latency (s) $\downarrow$} & \multicolumn{2}{c}{image-level} & \multicolumn{2}{c|}{scene-level}   & \multirow{2}{*}{Inf. Latency (s) $\downarrow$} \\ \cline{2-5} \cline{7-10}
& $\mathrm{AP}_{50} (\%) \downarrow$ & SSIM $\uparrow$ & $\mathrm{AP}_{50} (\%) \downarrow$ & \multicolumn{1}{c|}{SSIM $\uparrow$} &        & $\mathrm{AP}_{50} (\%) \downarrow$   & SSIM $\uparrow$  & $\mathrm{AP}_{50} (\%) \downarrow$ & \multicolumn{1}{c|}{SSIM $\uparrow$} &     \\ \hline
Normal &  98.3    &     - &   85.6       &   \multicolumn{1}{c|}{-}   &   -        &   97.8      &     -      &  91.4      & \multicolumn{1}{c|}{-}   & -  \\
Ours (SD v1.5)   &   18.3  & 0.972  &  16.6   &  \multicolumn{1}{c|}{0.837}     &    7.32   &   13.7    &   0.972     &    12.5  & \multicolumn{1}{c|}{0.840}     &  8.15  \\

Ours (SD v3.5)   & 17.0    &  0.978 &  27.9   &  \multicolumn{1}{c|}{0.845}     &    4.86    &     8.8   &   0.978     &    29.7  & \multicolumn{1}{c|}{0.844}     & 4.88  \\
\hline
\end{tabular}}   
\label{tab:supp-sd35}
\end{table}

\section{Extension to Rectified Flow}
\label{sec:supp_rectflow}
Recently, flow matching~\cite{flowmatching} has emerged as a powerful generative paradigm that learns a velocity field transporting a simple source distribution to the target data distribution. Among the variants built on flow matching,
Rectified flow~\cite{rectifiedflow} views the forward process as a linear interpolation between the latent $z_0$ and the noise $\varepsilon_t$, for $t \in [0,1]$:
\begin{equation}
\label{eqn:9}
    z_t = (1-t)z_0 + t\varepsilon_t, 
    \ \varepsilon_t \sim \mathcal{N}(0, \mathbf{I}), 
\end{equation}
The network $\epsilon_\theta (z_t, c)$ learns to estimate the velocity $(\varepsilon_t - z_0)$. 
Similarly, inspired by~\cite{dualimageprocess}, we adopt one-step estimate from noisy latent $z_t$ to approximate the reverse process:
\begin{equation}
\label{eqn:10}
    \hat{z_0} = z_t + t\epsilon_\theta (z_t, c)
\end{equation}
Similarly, we evaluate our pipeline using SD v3.5 \cite{stablediffusion3} as the underlying generative model and fine-tune a ControlNet \cite{ControlNet} in both stages with a batch size of 4. At inference time, we run the pipeline for 28 sampling steps, and all other settings follow the SD v1.5 configuration described in \cref{subsec:setup}. We conduct experiments on LINZ for the image-level strategy and on COCO for the scene-level strategy. Quantitative results comparing SD v1.5 and SD v3.5 are reported in \cref{tab:supp-sd35}. As shown in the table, adapting our pipeline to SD v3.5 yields performance comparable to SD v1.5 across all metrics, demonstrating that the attack remains effective, vehicle structure is well preserved, and the inference cost remains low. Qualitative comparisons on Faster-RCNN \cite{faster-rcnn}, presented in \cref{fig:supp_scene_level_sd35} and \cref{fig:supp_image_level_sd35}, further support these observations. As illustrated in the visualizations, SD v3.5 successfully transfers the appearance of  the reference areas while maintaining vehicle structure and reducing detector confidence, similar to SD v1.5.

\begin{table}
\caption{\textbf{Training parameters of each experiment}. ``lr'' denotes learning rate. ``iter'' denotes the total number of training iterations. ``$L_\text{struct}$, $L_\text{s}$, $L_\text{b}$, $L_\text{c}$, $L_\text{adv}$'' denotes the coefficents of structure preservation loss, style loss, background reconstruction loss, color-consistency loss, and adversarial loss, respectively.}

\resizebox{\textwidth}{!}{
\centering
\begin{tabular}{c|c|c|ccccc|ccccccc}
\hline
\multirow{2}{*}{Dataset} & \multirow{2}{*}{Target} & \multirow{2}{*}{Strategy} & \multicolumn{5}{c|}{No-Box Attack} & \multicolumn{7}{c}{White-Box Attack} \\ \cline{4-15} 
&  &  & $L_\text{struct}$   & $L_\text{s}$    &  $L_\text{b}$    & lr   & iter  & $L_\text{struct}$  & $L_\text{s}$  & $L_\text{b}$  & $L_\text{c}$  & $L_\text{adv}$  & lr  & iter  \\ \hline
COCO   & Faster-RCNN             &  Scene-Level  &  5.0    &  1.0    &   0.0   &     5e-6 &    12000    & 5.0   &  1.0  &   0.0 &   2.0 &  1.0  &   4e-6 &  15000     \\
COCO   & Faster-RCNN             &  Image-Level  &  2.5    &  2.5    &   1.0   &     2e-6 &    15000    &  4.0  & 1.0   &  1.0  &  2.0  &   1.0 &   2.5e-6  &  20000     \\
COCO   & ViTDet             & Scene-Level  & 5.0    &  1.0   &   0.0   &     5e-6 &    12000    &  5.0  &  1.0  & 0.0   &  2.0  & 1.0   &   4e-6  &  15000     \\
COCO   & ViTDet             & Image-Level  & 2.5    &  2.5   &   1.0   &     2e-6 &    15000    & 10.0   &  2.5  &  1.0  &  2.5  &  1.0  &   4e-6  &  20000     \\
\hline
LINZ  & Faster-RCNN    & Scene-Level         &  4.0    &  1.0    &   0.0   &     2.5e-6 &    10000    &  10.0  & 2.5    &  0.0  & 2.5   & 1.0   &   2e-6  &  12000     \\
LINZ  & Faster-RCNN    & Image-Level         &  5.0    &  5.0   &   1.0   &     2e-6 &    12000    &  10.0  &  7.5  &   1.0 &  7.5  &  1.0  &   2e-6  &  15000     \\
LINZ & ViTDet     &   Scene-Level &    4.0    &  1.0    &   0.0   &     2.5e-6 &    10000       &  10.0  & 2.5    &  0.0  & 2.5   & 1.0   &   2e-6  &  12000      \\
LINZ & ViTDet     &   Image-Level &    5.0    &  5.0   &   1.0   &     2e-6 &    12000  &  10.0  &  7.5  &   1.0 &  7.5  &  1.0  &   2e-6  &  15000      \\ \hline
\end{tabular}
}
\label{tab:supp-setup}
\end{table}

\section{Experiments}
\label{sec:supp_experiments}
\xiao{In this section, we describe more details regarding experiments. In~\cref{subsec:supp_setup}, we present more details regarding the training parameters and implementation details regarding other state-of-the-art methods. In~\cref{subsec:supp_qualitative}, We provide additional visualization results for both scene-level and image-level camouflage generation on both COCO and LINZ datasets. In~\cref{subsec:supp_scene_level}, we select another set of concepts in each scene to camouflage vehicles for the scene-level strategy to demonstrate our generalizability. In~\cref{subsec:supp_defense}, we evaluate the robustness of our camouflage attacks under common image preprocessing defense strategies. In~\cref{subsec:supp_ablation},  we present more ablation studies regarding the effectiveness of our two-stage pipeline against a one-stage variant, structure preservation loss $L_\text{struct}$, style loss $L_\text{s}$, and adversarial loss $L_
\text{adv}$.}

\subsection{Experimental Setup}
\label{subsec:supp_setup}
In~\cref{subsec:setup}, we discuss common training parameters across all experiments, such as training batch size and the dilation kernel size of COCO and LINZ. In this section, we summarize other parameters, including learning rate, total number of training iterations, and coefficients of loss functions, as listed in~\cref{tab:supp-setup}.
Empirically, we observe that setting a higher coefficient for other loss functions than adversarial loss is beneficial to set higher constraint to the style and structure preservation while achieving high attack effectiveness. During inference, camouflaged images are sampled one at a time and all methods are tested on a single RTX A6000 GPU.

Moreover, we provide implementation details for the baseline methods listed in \cref{tab:sota-coco} and \cref{tab:sota-other-types}, grouped into three categories as described in~\cref{sec:related_work}: \textit{imperceptible perturbations}, \textit{adversarial patches}, and \textit{ stylization attacks}.

\begin{itemize}
    \item \textit{Imperceptible perturbations}: TOG~\cite{TOG} directly injects bounded RGB-space perturbations without requiring any reference style, and we constrain the perturbation norm to $\frac{100}{255}$ with 80 optimization iterations per image. DiffAttack~\cite{diffattack}, in contrast, performs optimization in the latent space of a diffusion model to preserve perceptual fidelity. We use a 50-step diffusion process and 60 adversarial optimization steps. Since DiffAttack was originally designed for image classification, we adapt it to object detection by replacing its adversarial loss with our detector-specific loss in Eq.~\ref{eq:adv_loss}.
    
    \item \textit{Adversarial patches}: For NAP~\cite{NAP}, we optimize a GAN-generated patch and use a single shared patch for all images. Following the original implementation, we train the patch on the LINZ and COCO training sets for 100 epochs with a learning rate of 1e-4, then apply the optimized patch to the test sets. The patch is scaled to remain fully within the object region. BadPatch~\cite{badpatch} follows a similar pipeline but initializes from a natural image patch and optimizes it in the noisy diffusion latent space via DDIM inversion. We use 25 inversion steps and train the patch for 20 epochs with a learning rate of 1e-4 on both datasets before evaluating on the test sets.
    \item \textit{Stylization attacks}: Both AdvCAM~\cite{AdvCam} and Diff-PGD~\cite{DiffPGD} require a reference patch to guide stylization. Because their pipelines only accept square patches, we adopt a unified reference-selection strategy: for the image-level setting, we choose a square patch near the target vehicle; for the scene-level setting, we crop a square patch from the scene’s reference area. Since both methods were originally designed for classifier attacks, we replace their adversarial losses with our detector-specific loss in Eq.~\ref{eq:adv_loss}. For each image, we perform 1,000 optimization iterations to obtain the final stylized results.
\end{itemize}

\begin{figure}[!tb]
    \centering  
    \includegraphics[width=0.75\columnwidth]{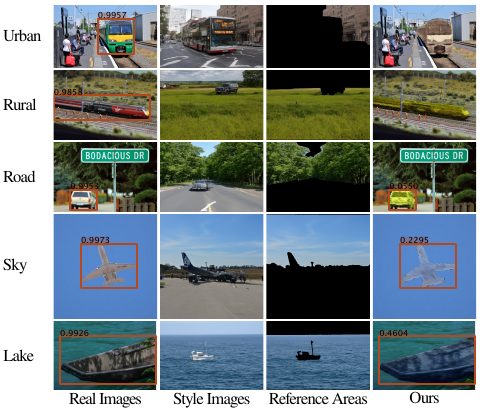}
    \caption{\textbf{More Visualizations on the COCO dataset.} We visualize scene-level camouflage generation results.  We present examples in  \textit{urban, rural, road, sky}, and \textit{lake} scenes, where they are associated with 
 \textit{building, grass, tree, sky}, and \textit{water} as representative concepts.
    }
    \label{fig:supp_coco_scene_level}
\end{figure}

\begin{figure}[!tb]
    \centering  
    \includegraphics[width=0.75\columnwidth]{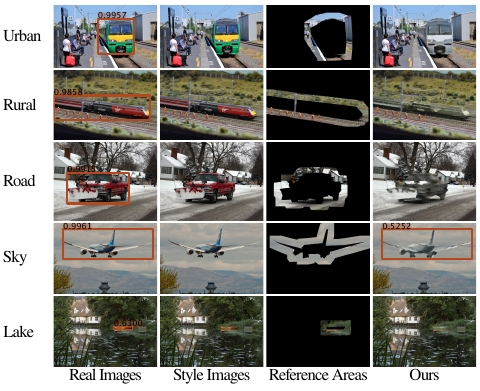}
    \caption{\textbf{More Visualizations on the COCO dataset.} We visualize image-level camouflage generation results. 
    }
    \label{fig:supp_coco_image_level}
\end{figure}

\begin{figure}[!tb]
    \centering  
    \includegraphics[width=0.75\columnwidth]{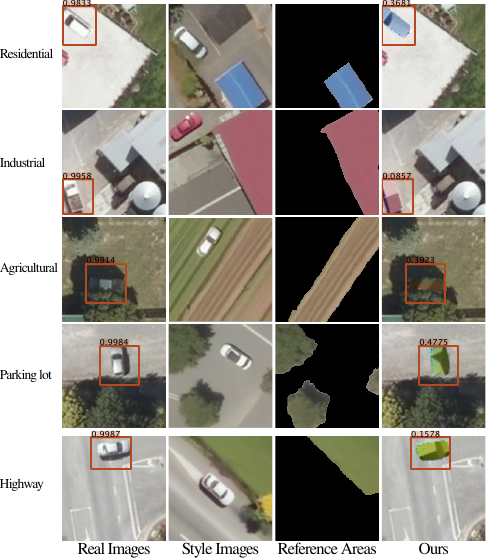}
    \caption{\textbf{Visualizations on the LINZ dataset.} We present examples in \textit{residential, industrial, agricultural, parking lot,} and \textit{highway} scenes, where they are matched with \textit{house, building, field, tree}, and \textit{grass} as representative concepts.
    }
    \label{fig:supp_linz_scene_level}
\end{figure}

\begin{figure}[!tb]
    \centering  
    \includegraphics[width=0.75\columnwidth]{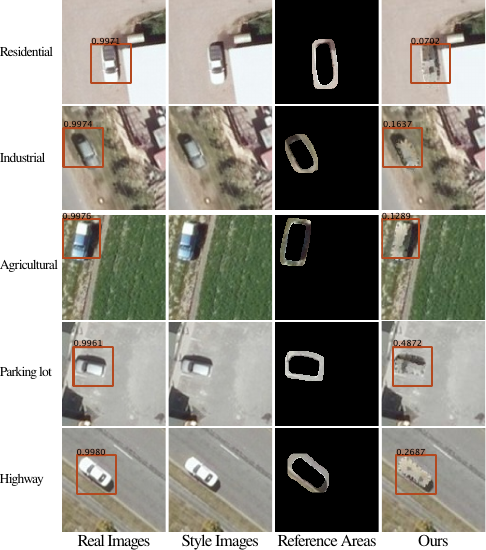}
    \caption{\textbf{More Visualizations on the LINZ dataset.} We visualize image-level camouflage generation results. 
    }
    \label{fig:supp_linz_image_level}
\end{figure}

\xiao{\subsection{Extended Qualitative Results}
\label{subsec:supp_qualitative}
We provide additional visualization results to complement~\cref{fig:sota}, covering both scene-level and image-level camouflage generation on the COCO dataset in~\cref{fig:supp_coco_scene_level} and~\cref{fig:supp_coco_image_level}. Scene-level examples span \textit{urban, rural, road, sky}, and \textit{lake} environments paired with visual concepts such as \textit{building, grass, tree, sky}, and \textit{water}. Image-level examples use reference areas extracted from the vehicle’s immediate surroundings. Across both settings, the camouflaged vehicles blend naturally with the scene while retaining their geometric structure, demonstrating consistent stylization behavior.

Additional visualizations for LINZ are shown in~\cref{fig:supp_linz_scene_level} and~\cref{fig:supp_linz_image_level}, covering \textit{residential, industrial, agricultural, parking lot}, and \textit{highway} environments. Scene-level stylization follows representative concepts like \textit{house, building, field, tree}, and \textit{grass}, while image-level stylization leverages local surroundings. In both cases, our method produces camouflaged vehicles that maintain realistic structure and visual coherence with their surroundings.

Together, these visualizations corroborate that the proposed camouflage generation approach generalizes across datasets and stylization modes, consistently producing structurally faithful and stealthy camouflage.}

\setlength{\tabcolsep}{3pt}
\begin{table}[!tb]
\caption{\textbf{Quantitative evaluation of alternative scene–concept pairings for the scene-level strategy on COCO and LINZ.} ``Ours'' denotes the scene–concept groups described in~\cref{subsec:setup}, while ``Ours*'' adopts the alternative set evaluated in~\cref{subsec:supp_scene_level}.
}

\centering
\small
\begin{tabular}{c|c|cc|cc}
\hline
\multirow{2}{*}{Dataset} & \multirow{2}{*}{Method} & \multicolumn{2}{c|}{Faster-RCNN} & \multicolumn{2}{c}{ViTDet} \\ \cline{3-6} 
&                         & $\mathrm{AP}_{50} (\%) \downarrow$ & SSIM $\uparrow$ & $\mathrm{AP}_{50} (\%) \downarrow$ & SSIM   $\uparrow$     \\ \hline
COCO        & Normal    & 85.6                &   -             &    91.4          &  -           \\
COCO & Ours                    &         16.6        &       0.837         &   12.5           &   0.840          \\
COCO                     & Ours*    &   21.1              &     0.846           &         20.2     &  0.845           \\ \hline
LINZ &    Normal                    & 98.3                &       -         &     97.8         &    -         \\
LINZ   &     Ours      &       27.5          &        0.961        &  11.1            &  0.964           \\
LINZ &     Ours*     &       24.8          &           0.966     &   14.5           &    0.969    \\ \hline
\end{tabular}

\label{tab:supp-scene-group2}
\end{table}

\begin{figure}[!tb]
    \centering  
    \includegraphics[width=0.75\columnwidth]{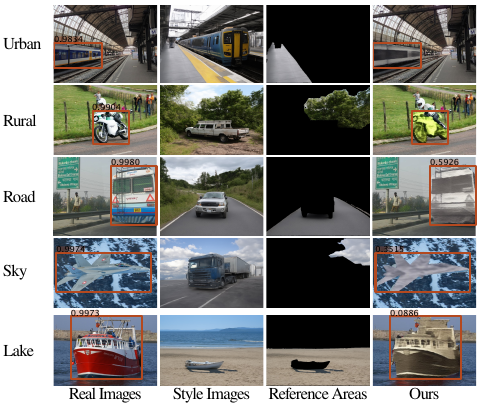}
    \caption{\textbf{Visualizations on the COCO dataset.} We present examples in \textit{urban, rural, road, sky}, and \textit{lake} scenes, where they are matched with 
 with \textit{platform, tree, road, cloud}, and \textit{beach} as representative concepts.
    }
    \label{fig:supp_coco_scene_level_group2}
\end{figure}

\begin{figure}[!tb]
    \centering  
    \includegraphics[width=0.75\columnwidth]{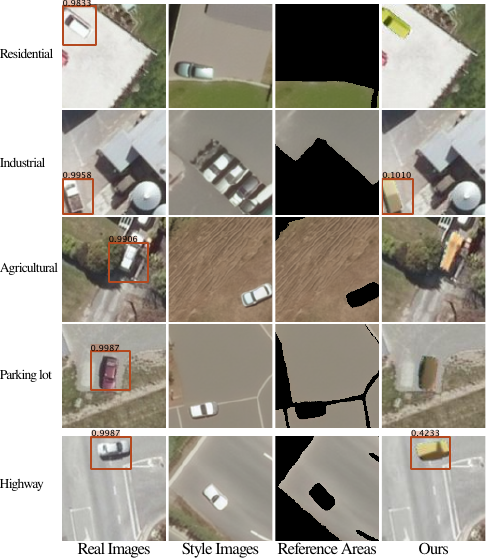}
    \caption{\textbf{Visualizations on the LINZ dataset.} We present examples in \textit{residential, industrial, agricultural, parking lot,} and \textit{highway} scenes, where they are matched with 
 with \textit{grass, sidewalk, parking lot, dirt}, and \textit{parking lot} as representative concepts.
    }
    \label{fig:supp_linz_scene_level_group2}
\end{figure}

\subsection{Scene-Level Strategy}
\label{subsec:supp_scene_level}
As described in~\cref{sec:intro}, the scene-level strategy adapts the vehicle's appearance to a common semantic concept drawn from the broader scene, ensuring that the resulting camouflage remains visually coherent and avoids salient or unnatural patterns that may draw human attention. In \cref{subsec:sota}, we evaluate one set of scene–concept pairings for both datasets. 
To assess the generalizability of this strategy, we conduct an additional experiment on both datasets using an alternative set of scene and concept pairings.

Following~\cref{subsec:setup}, COCO images are assigned to five environments: \textit{urban, rural, road, sky}, and \textit{lake}. In~\cref{sec:experiments}, these environments are paired with  \textit{building, grass, tree, sky}, and \textit{water} as representative style concepts. In this section, we additionally evaluate an alternative set of pairings that matches the same environments 
 with \textit{platform, tree, road, cloud}, and \textit{beach}.  Similarly, LINZ images are assigned \textit{residential, industrial, agricultural, parking lot}, and \textit{highway} scenes, originally associated with  \textit{house, building, field, tree}, and \textit{grass}.  In this section, we also evaluate an alternative set of pairings that matches these environments with \textit{grass, sidewalk, parking lot, dirt}, and \textit{parking lot}.  
 
 The quantitative results in~\cref{tab:supp-scene-group2} demonstrate that both pairing sets achieve comparable attack effectiveness and preservation of vehicle structure for both datasets. Additional qualitative examples in \cref{fig:supp_coco_scene_level_group2} and \cref{fig:supp_linz_scene_level_group2} further confirm this consistency. Overall, adapting vehicles to scene-relevant visual concepts produces coherent and stealthy camouflage, though occasional style misalignment between camouflaged vehicles and reference areas occurs, which is further discussed in \cref{sec:supp_limitations}.

\begin{table}[!tb]
\caption{\textbf{Evaluation under preprocessing-based defense strategies on the COCO dataset.} ``Normal'' denotes clean images, ``Baseline'' denotes adversarial images without defense, ``Denoise'' applies Non-local Means denoising, and ``Smooth'' applies bilateral filtering.}
\centering
\footnotesize
\begin{tabular}{c|cc|cc}
\hline
\multirow{2}{*}{Method} & \multicolumn{2}{c|}{Faster-RCNN} & \multicolumn{2}{c}{ViTDet} \\  
& image-level            & scene-level & image-level        & scene-level     \\ \hline
Normal      &   85.6    &  85.6 &  91.4  &   91.4    \\
Baseline      &   15.0    &  16.6 &  19.2  &   12.5   \\
Denoise                  &   29.8     &   28.2        & 36.4  &  27.4     \\
Smooth                   &   21.4     &  25.4     & 25.5   &  23.0  \\ 
\hline
\end{tabular}
\label{tab:supp-defense}
\end{table}
\xiao{\subsection{Robustness under defense strategies}
\label{subsec:supp_defense}
We evaluate the robustness of our camouflage attacks under common image preprocessing defenses, including Non-local Means denoising and bilateral filtering, on adversarial images from COCO. As shown in~\cref{tab:supp-defense}, these defenses weaken attack effectiveness. However, our method remains highly destructive. Across both Faster R-CNN and ViTDet, and under both image-level and scene-level attacks, we consistently observe large $\mathrm{AP}_{50}$ drop
 relative to clean images. For example, $\mathrm{AP}_{50}$
decreases from 85.6\% to 29.8\% under denoising for Faster R-CNN, and from 91.4\% to 36.4\% for ViTDet using image-level strategy. Under smoothing, $\mathrm{AP}_{50}$
remains below 26\% in all cases. These results indicate that our attacks are not easily removed by standard denoising or smoothing defenses. We attribute this robustness to the fact that our method modifies object appearance rather than relying on fragile high-frequency perturbations, making it inherently more resistant to such preprocessing methods.
}

\begin{figure}[!tb]
    \centering  \includegraphics[width=0.65\columnwidth]{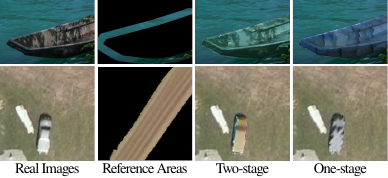}
    \caption{\textbf{Comparing our pipeline with one-stage variants}.}
    \label{fig:two_stage}
\end{figure}
\setlength{\tabcolsep}{2.5pt}
\begin{table}[!tb]
\caption{\textbf{Effectiveness of adversarial loss.}}
\centering
\footnotesize
\begin{tabular}{c|c|cc|cc}
\hline
\multirow{2}{*}{Dataset} & \multirow{2}{*}{Method} & \multicolumn{2}{c|}{Faster-RCNN} & \multicolumn{2}{c}{ViTDet} \\ \cline{3-6} &                         & image-level     & scene-level    & image-level  & scene-level \\ 
\hline
COCO  & Normal & 85.6 & 85.6 & 91.4 & 91.4 \\
COCO    &  W/O $L_\text{adv}$       &     67.2       &  65.1              &     73.6         &  72.1           \\
COCO    &  Ours       &    18.3             &    27.5            &    13.7          &     11.1        \\
\hline
LINZ & Normal & 98.3 & 98.3 & 97.8 & 97.8 \\
LINZ      &    W/O $L_\text{adv}$       &  92.4        &        77.2        &  85.2            &   52.9     \\
LINZ     &   Ours   &  15.0     &  16.6   &  19.2   &  12.5          \\ 
\hline
\end{tabular}
\label{tab:supp-abl-adv}
\end{table}

\begin{figure}[!tb]
    \centering      \includegraphics[width=0.65\columnwidth]{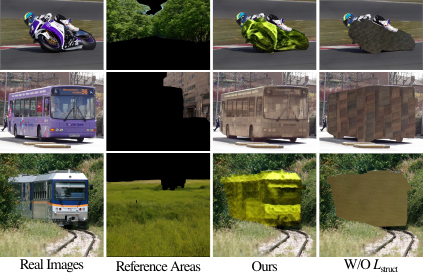}
    \caption{\textbf{Effectiveness of structure preservation loss.}
    }
    \label{fig:supp_structure loss}
\end{figure}

\begin{figure}[!tb]
    \centering  \includegraphics[width=0.65\columnwidth]{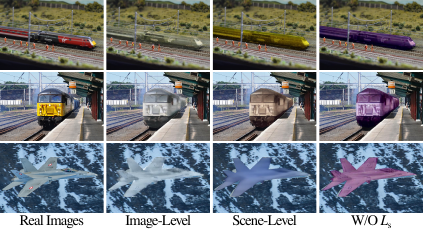}
    \caption{\textbf{Effectiveness of style loss.}
    }
    \label{fig:supp_style loss}
\end{figure}

\subsection{Ablation Studies}
\label{subsec:supp_ablation}

\xiao{\noindent\textbf{Two-stage vs. One-stage.} We evaluate the effectiveness of our proposed two-stage pipeline against a one-stage variant that jointly fine-tunes the ControlNet using the structure preservation loss $L_\text{struct}$, style loss $L_\text{s}$, adversarial loss $L_\text{adv}$, and background supervision loss $L_\text{b}$ under the image-level setting, as illustrated in~\cref{fig:two_stage}. The two-stage pipeline achieves more accurate stylization of vehicles according to the visual characteristics of the reference area. This improvement arises because, in the early phase of training, generative pipelines often fail to reconstruct the vehicle at its original location via~\cref{eqn:2}, producing arbitrary content instead. Consequently, the adversarial loss $L_\text{adv}$ becomes misleading, since detectors already assign low confidence to non-vehicle regions, which does not reflect true adversarial success. This inaccurate supervision interferes with the optimization of other objectives and leads to unstable training. By first ensuring reliable vehicle reconstruction before applying adversarial learning, the two-stage design stabilizes optimization and produces higher-quality camouflage.}

\noindent\textbf{Effectiveness of structure preservation loss.} We evaluate the effectiveness of structure preservation loss $L_\text{struct}$  for the scene-level strategy on the COCO dataset, as shown in~\cref{fig:supp_structure loss}. When the structure-preservation loss is removed, the model fails to maintain the geometry of the vehicle and instead produces heavily distorted shapes that no longer resemble the underlying object. Although the stylization remains roughly consistent with the reference area, the resulting camouflages become unrealistic, demonstrating that $L_\text{struct}$ is essential for preserving vehicle structure while adapting appearance.

\noindent\textbf{Effectiveness of style loss.} We evaluate the effectiveness of style loss $L_\text{s}$ on the COCO dataset, as illustrated in~\cref{fig:supp_style loss}. When the style loss is removed, the pipeline tends to learn a similar stylization for all vehicles, often producing colors or textures that appear unnatural and may draw human attention. In contrast, by guiding the model to transfer each vehicle’s appearance either to its surrounding region in the image level setting or to an appropriate visual concept present in the scene in the scene level setting, the style loss encourages camouflages that remain consistent with the environment.

\noindent\textbf{Effectiveness of adversarial loss.} We evaluate the effectiveness of the adversarial loss $L_\text{adv}$ introduced in~\cref{subsec:white box attack}. Removing this term is equivalent to performing only the No-Box Attack stage described in~\cref{subsec:no-box attack}. Admittedly, style-based editing alone reduces detector confidence by altering the vehicle appearance, as shown in~\cref{tab:supp-abl-adv}.  However, our stylization procedures are not designed to introduce patterns that deviate significantly from the distribution of the dataset. Consequently, incorporating the adversarial loss further decreases detector confidence by explicitly optimizing for adversarial behavior. 

\begin{figure}[!tb]
    \centering  
    \includegraphics[width=0.98\columnwidth]{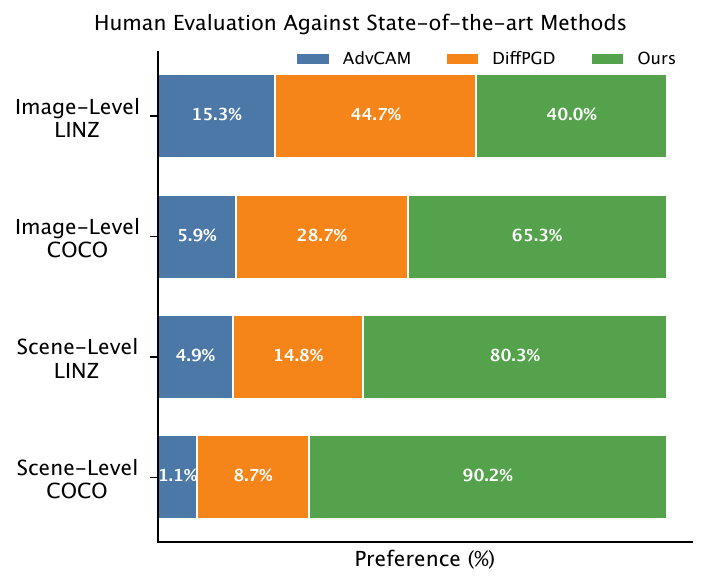}
    \caption{\textbf{Human Evaluation of stealthiness comparing with state-of-the-art methods.}}
    \label{fig:supp_questionnaire_compare_with_sota}
\end{figure}

\begin{figure}[!tb]
    \centering  
    \includegraphics[width=0.75\columnwidth]{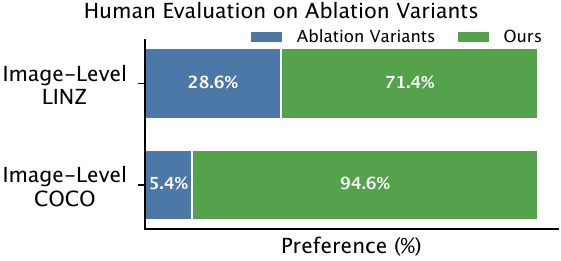}
    \caption{\textbf{Human Evaluation of stealthiness across ablation variants.}}
    \label{fig:supp_questionnaire_ablation}
\end{figure}

\begin{figure}[!tb]
    \centering  
    \includegraphics[width=0.98\columnwidth]{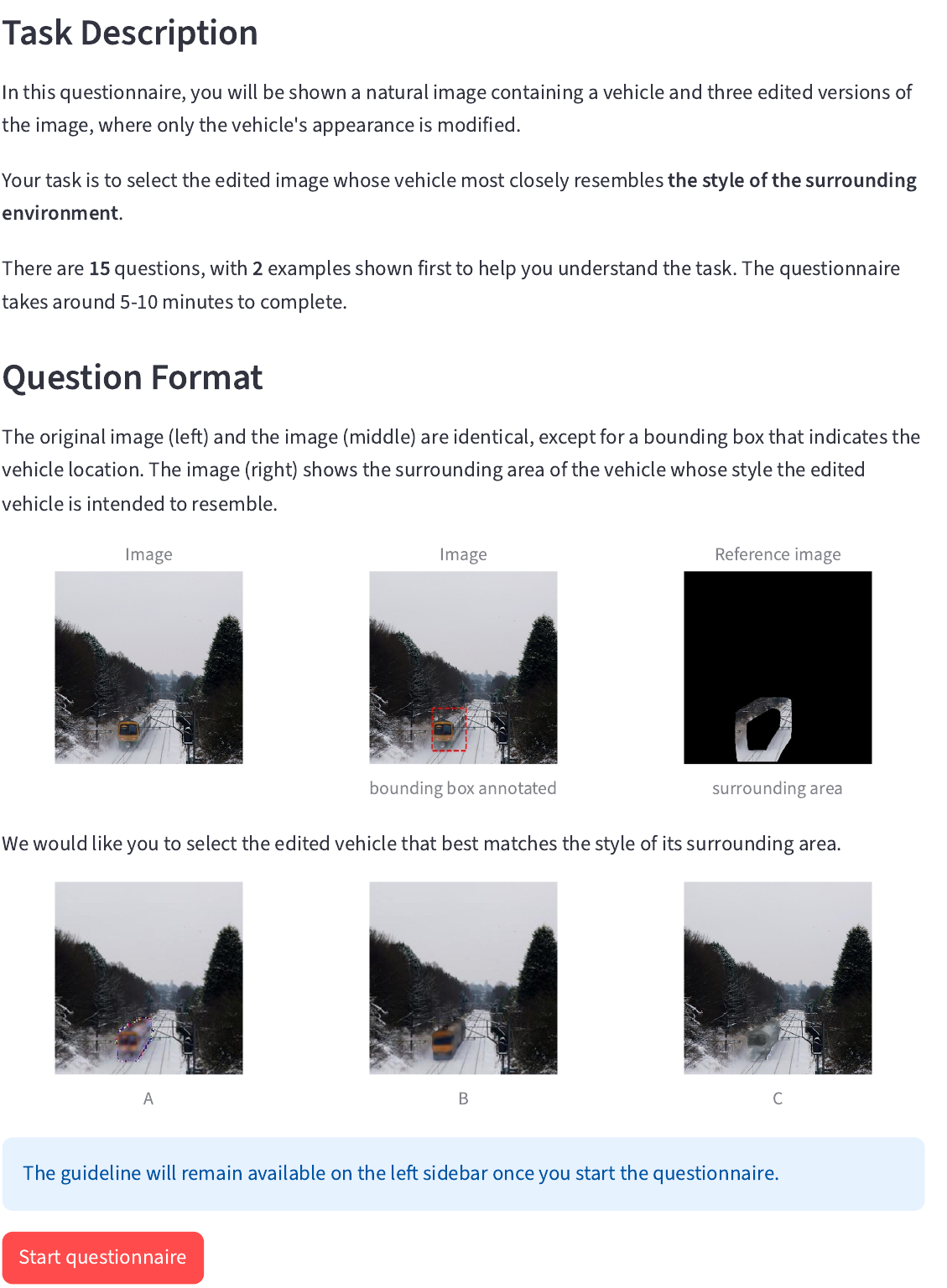}
    \caption{\textbf{Image-level user study guideline.}}
    \label{fig:supp_guideline_image_level}
\end{figure}

\begin{figure}[!tb]
    \centering  
    \includegraphics[width=0.75\columnwidth]{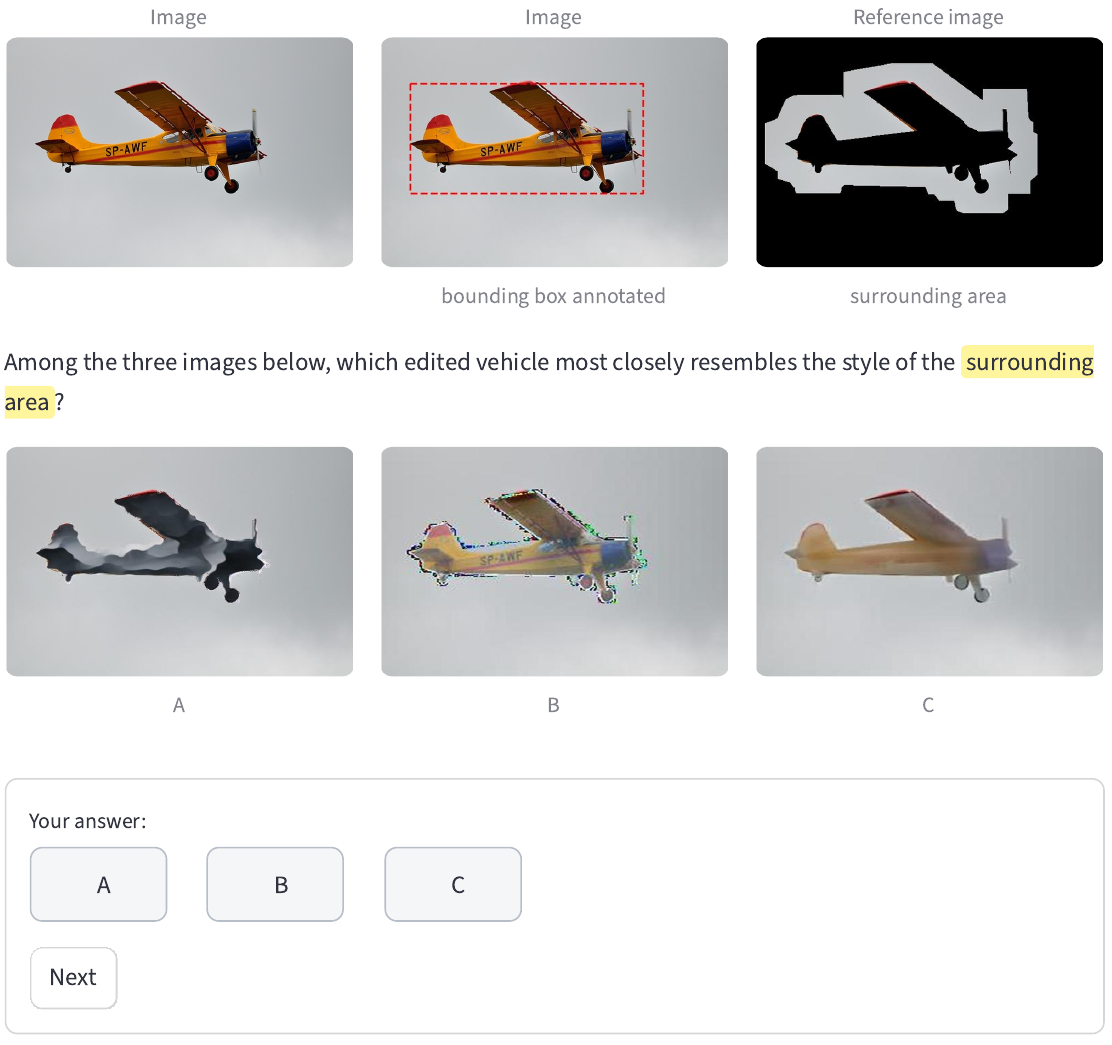}
    \caption{\textbf{Image-level example question on the COCO dataset.}}
    \label{fig:supp_question_coco_image_level}
\end{figure}

\begin{figure}[!tb]
    \centering  
    \includegraphics[width=0.75\columnwidth]{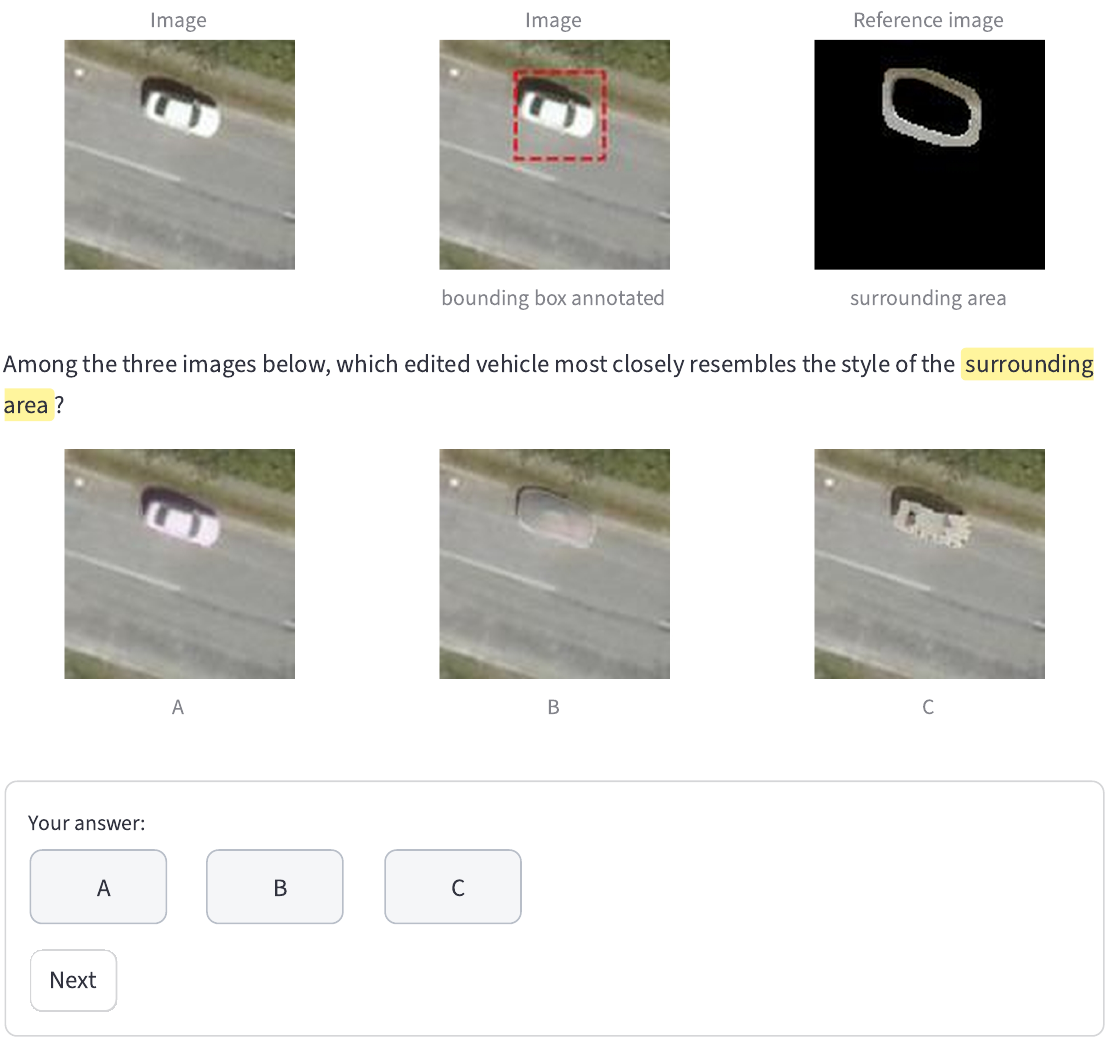}
    \caption{\textbf{Image-level example question on the LINZ dataset.}}
    \label{fig:supp_question_linz_image_level}
\end{figure}

\begin{figure}[!tb]
    \centering  
    \includegraphics[width=0.98\columnwidth]{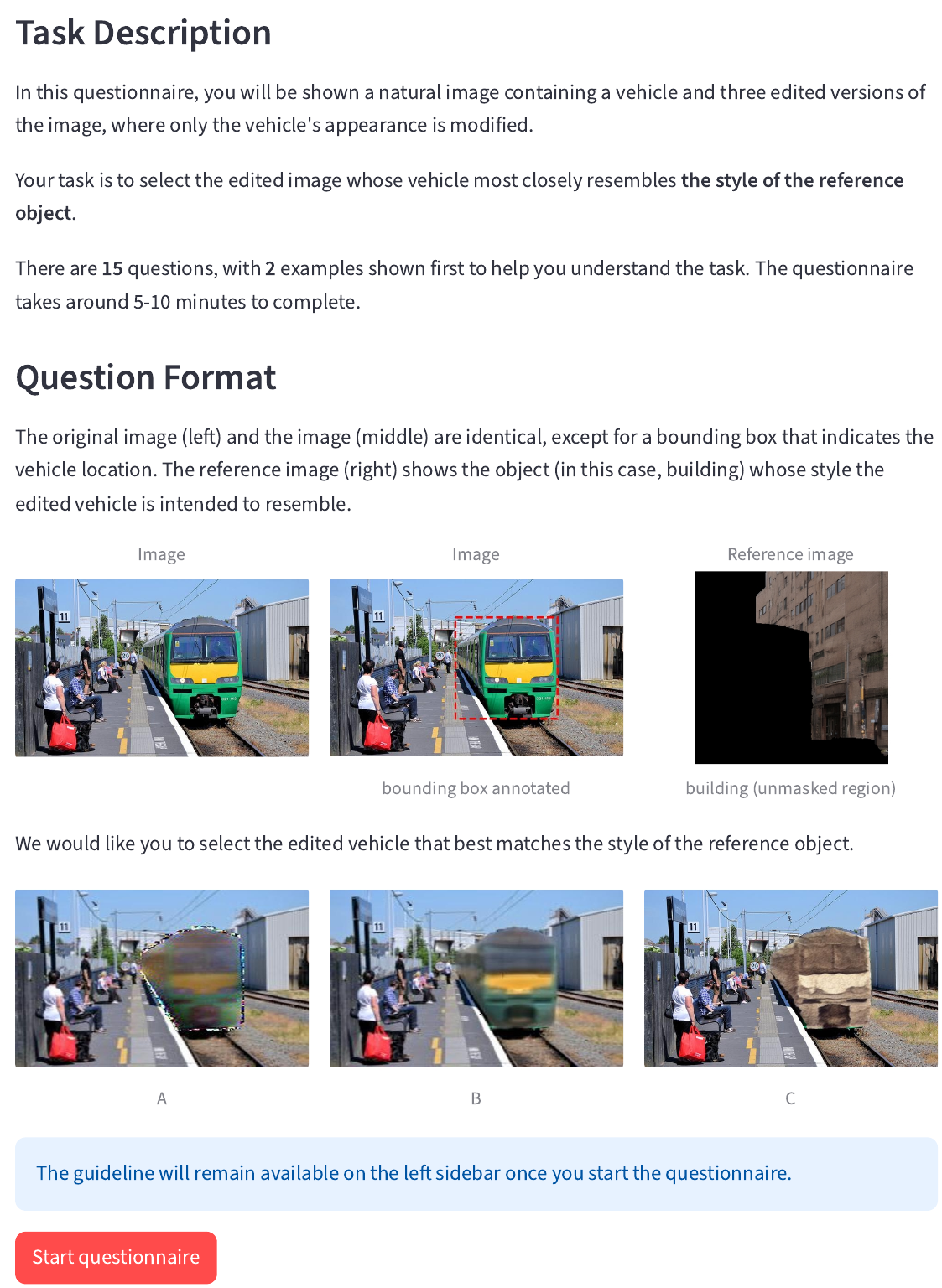}
    \caption{\textbf{Scene-level user study guideline.}}
    \label{fig:supp_guideline_scene_level}
\end{figure}

\begin{figure}[!tb]
    \centering  
    \includegraphics[width=0.75\columnwidth]{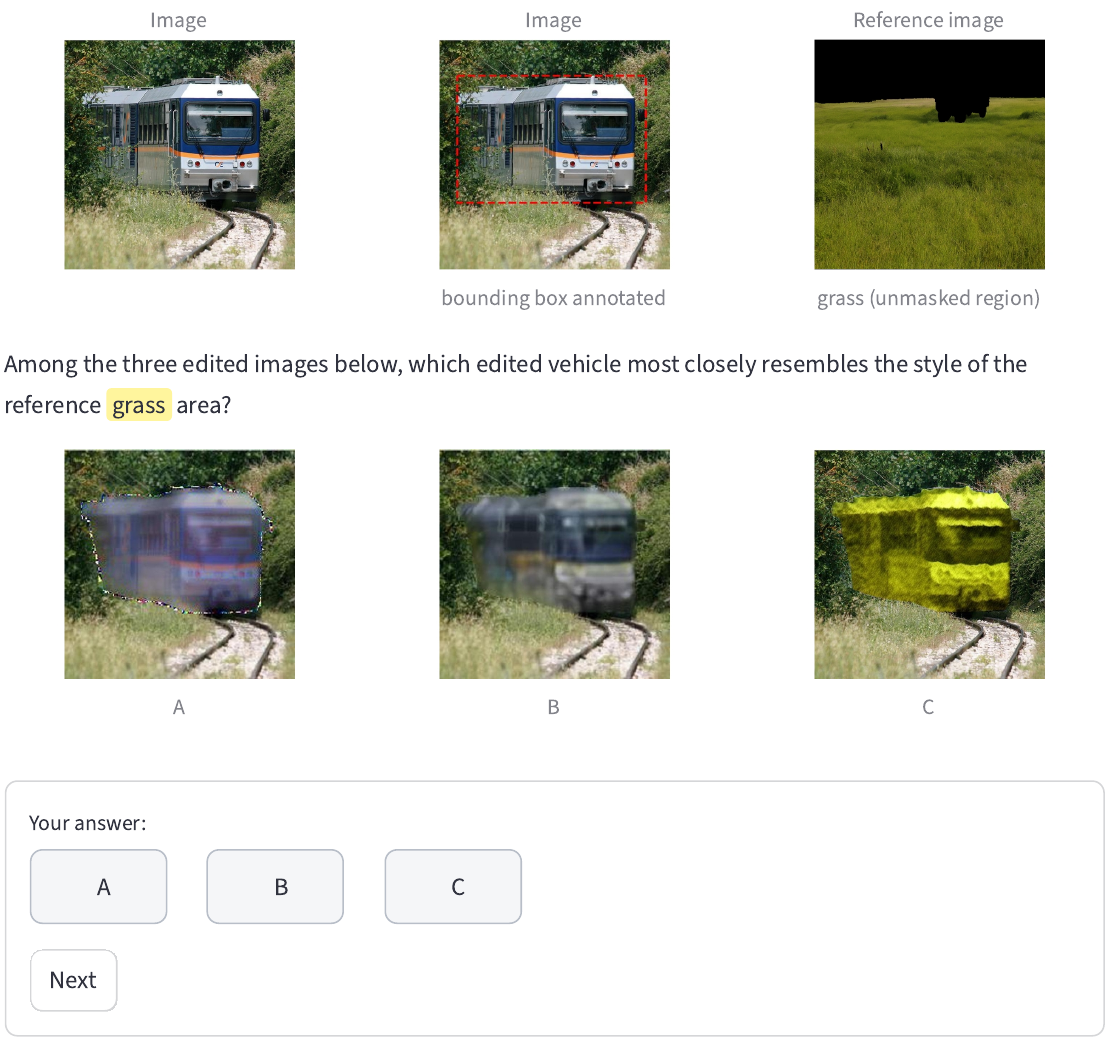}
    \caption{\textbf{Scene-level example question on the COCO dataset.}}
    \label{fig:supp_question_coco_scene_level}
\end{figure}

\begin{figure}[!tb]
    \centering  
    \includegraphics[width=0.75\columnwidth]{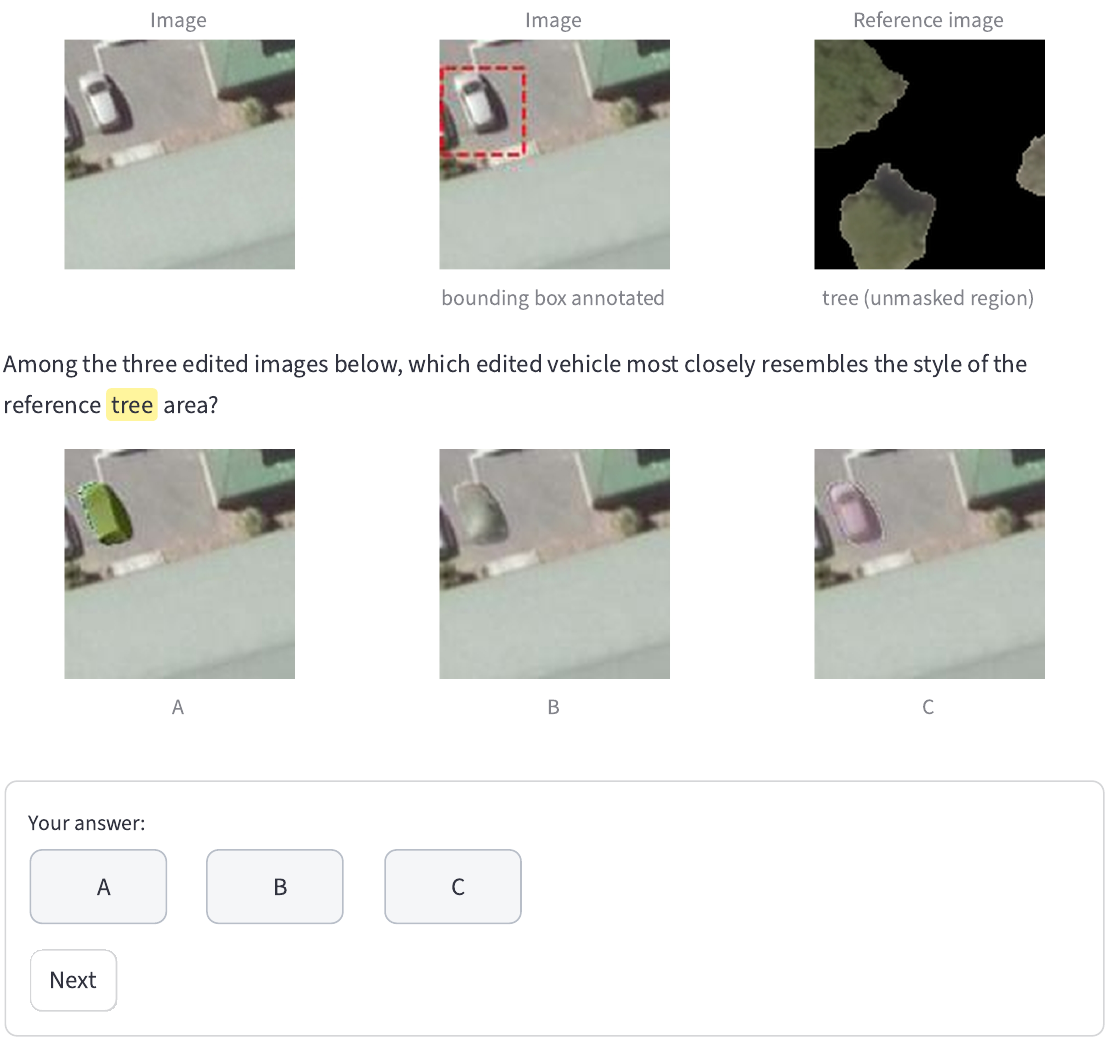}
    \caption{\textbf{Scene-level example question on the LINZ dataset.}}
    \label{fig:supp_question_linz_scene_level}
\end{figure}

\begin{figure}[!tb]
    \centering  
    \includegraphics[width=0.75\columnwidth]{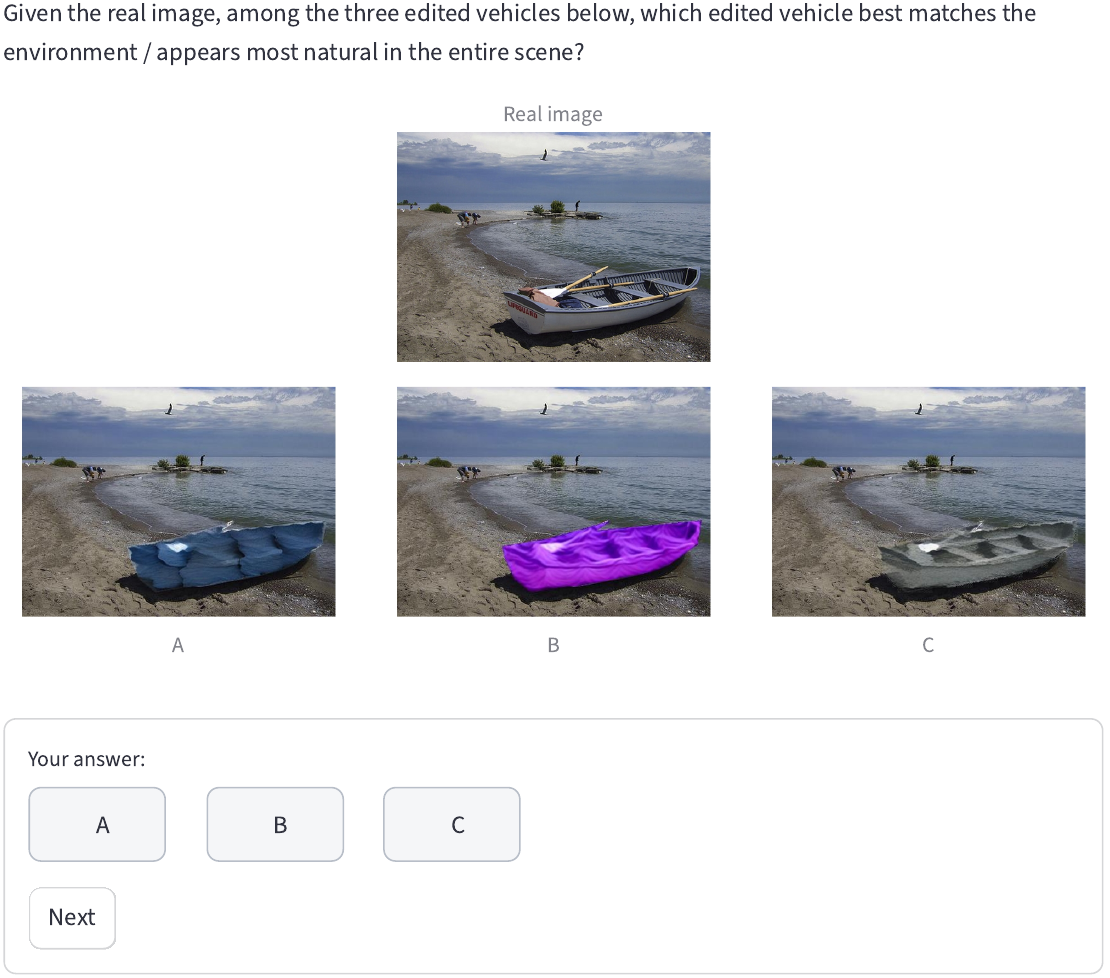}
    \caption{\textbf{Example question on the COCO dataset.}}
    \label{fig:supp_question_coco_ablation}
\end{figure}

\begin{figure}[!tb]
    \centering  
    \includegraphics[width=0.75\columnwidth]{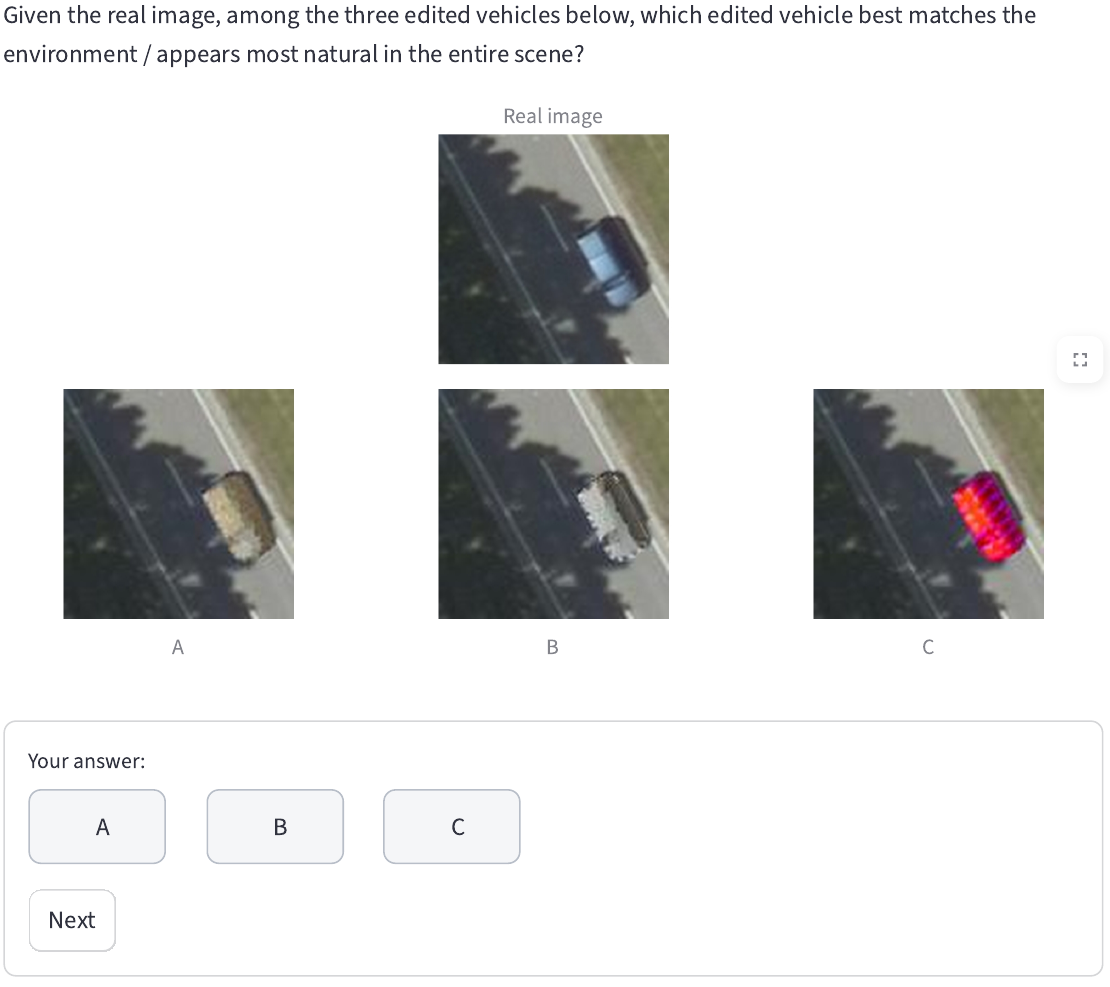}
    \caption{\textbf{Example question on the LINZ dataset.}}
    \label{fig:supp_question_linz_ablation}
\end{figure}

\xiao{\section{Human Evaluation}
\label{sec:supp_human}
In this section, we provide additional details on the evaluation on stealthiness via human studies. In~\cref{subsec:supp_human_study_compare_sota}, we compare with state-of-the-art approaches to evaluate stealthiness via preferences on whose stylization best matched the reference areas. In~\cref{subsec:supp_human_study_ablation}, we compare with state-of-the-art approaches to evaluate stealthiness via preferences on the naturalness of edited vehicles.

\subsection{Comparison with State-of-the-art Methods}
\label{subsec:supp_human_study_compare_sota}
We provide additional details for the user study described in~\cref{subsec:sota}, which evaluates the perceived stealthiness of our method compared with state-of-the-art approaches. The questionnaire is divided into two parts corresponding to the two camouflage strategies: image-level and scene-level. The order of these two parts is randomized for each participant. The question pool contains 15 images from the LINZ dataset and 15 images from COCO. There are three images in each scene type. For each image, camouflaged results are generated using both image-level and scene-level strategies for all compared methods. In each part, participants are first presented with a guideline explaining the evaluation task, followed by 15 questions randomly sampled from the pool. Participants are asked to select the method whose stylization best matched the reference areas. The guidelines for the two parts are shown in~\cref{fig:supp_guideline_image_level} and~\cref{fig:supp_guideline_scene_level}. Example questions for both datasets and both camouflage strategies are shown in~\cref{fig:supp_question_coco_image_level},~\cref{fig:supp_question_coco_scene_level},~\cref{fig:supp_question_linz_image_level}, and~\cref{fig:supp_question_linz_scene_level}. The aggregated preference results are shown in~\cref{fig:supp_questionnaire_compare_with_sota}. On the LINZ dataset under the image-level strategy, our method achieves a preference rate comparable to Diff-PGD~\cite{DiffPGD}. In contrast, for the other settings, including image-level on COCO and both scene-level strategies, our method receives substantially higher preference rates than competing approaches. These results indicate that our method achieves improved stylization quality and produces camouflage patterns that are consistently perceived by human evaluators as more stealthy.

\subsection{Comparison with Ablation Variants}
\label{subsec:supp_human_study_ablation}
In addition to the comparison with state-of-the-art methods, we conduct a separate human study to analyze the effect of different design choices through ablation variants. Directly evaluating stealthiness against previous methods is challenging because their outputs may differ in multiple aspects beyond appearance style, such as structural preservation and adversarial effectiveness (see~\cref{tab:sota-linz} and~\cref{tab:sota-coco}). These factors can influence human perception of style realism and therefore confound the evaluation of stealthiness~\cite{PhysicalAttackNaturalness}. To mitigate this issue, we design a human study focusing on ablation variants of our image-level strategy by selectively removing key loss terms. This setting ensures that all edited vehicles maintain comparable structural consistency and adversarial effectiveness through the use of the structure-preservation loss and adversarial loss, so that the primary variation lies in the resulting appearance style. For the COCO dataset, we compare against variants that remove the style loss or the background reconstruction loss. For the LINZ dataset, we compare against variants that remove the style loss and use the single-stage generation paradigm.
We use the same image pool as in~\cref{subsec:supp_human_study_compare_sota}. Participants are first provided with a guideline explaining the evaluation task, followed by 10 questions randomly sampled from the image pool. In each question, participants are shown the real image together with three edited vehicles and asked to select the version that appears most natural within the scene context. Example questions are illustrated in~\cref{fig:supp_question_linz_ablation} and~\cref{fig:supp_question_coco_ablation}.The aggregated results are summarized in~\cref{fig:supp_questionnaire_ablation}. Our method achieves substantially higher preference rates than other ablation variants on both datasets, receiving 71.4\% preference on LINZ and 94.6\% on COCO. These results indicate that the complete formulation of our method produces camouflage patterns that better match the surrounding environment and are consistently perceived as more natural and stealthy by human evaluators.  
}

\begin{figure}[!tb]
    \centering  
    \includegraphics[width=0.65\columnwidth]{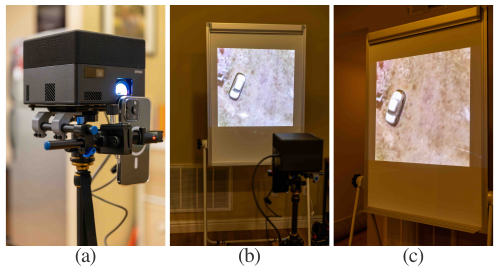}
    \caption{\textbf{Projector-based physical setup for the LINZ dataset.} (a) Device configuration with the projector and camera aligned to minimize parallax. (b) A real image is projected onto a whiteboard with a 3D-printed car model placed at the corresponding location. (c) Alternative view showing attachment of the printed model to the board.
    }
    \label{fig:supp_projector_linz_setup}
\end{figure}

\begin{figure}[!tb]
    \centering  
    \includegraphics[width=0.85\columnwidth]{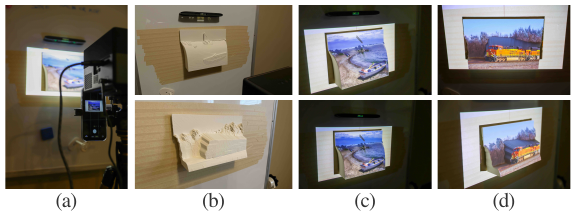}
    \caption{\xiao{\textbf{Projector-based physical setup for the COCO dataset.} (a) Hardware configuration. (b) 3D-printed scene reconstructed from two COCO images via single-view reconstruction. (c-d) Real images projected onto the corresponding 3D-printed models.
    }}
    \label{fig:supp_projector_coco_setup}
\end{figure}

\begin{figure}[!tb]
    \centering  \includegraphics[width=0.98\columnwidth]{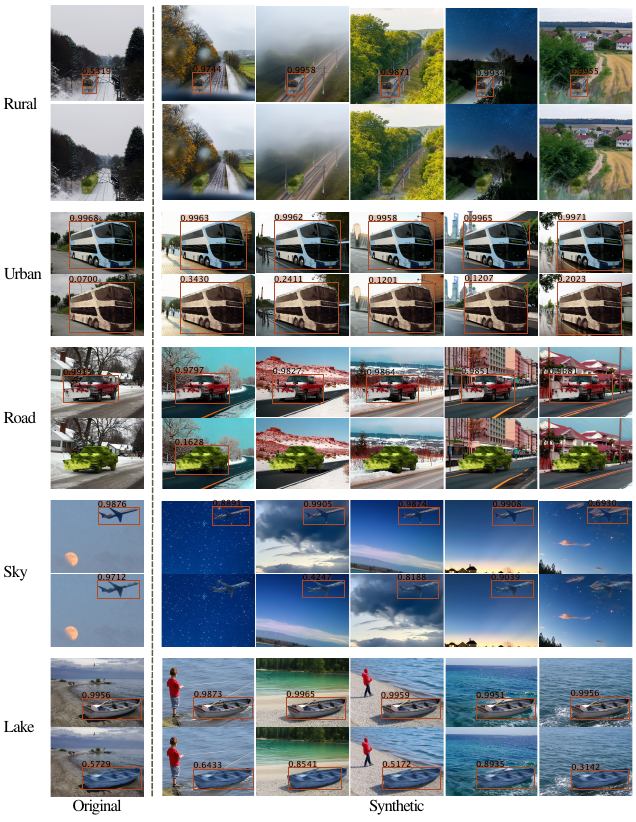}
    \caption{\textbf{Transferability of scene-level camouflage across locations in COCO dataset.} For each scene type, we composite the vehicle into five additional backgrounds to create synthetic scenes while keeping the vehicle appearance unchanged. In each scene group, the first row shows detection results on clean vehicles, and the second row shows results on camouflaged vehicles. “Original” denotes the original COCO scene, whereas “Synthetic” refers to the newly composed scenes with replaced backgrounds.
    }
    \label{fig:supp_transferable_other_places}
\end{figure}

\xiao{\section{Transferability}
\label{sec:supp_transferability}
In this section, we provide additional details on the transferability robustness of our method. In~\cref{subsec:supp_cross_location_transferability}, we conduct an additional experiment to evaluate the transferability of scene-level camouflage across different locations within the same scene category. In~\cref{subsec:supp_physical_experiment}, we present additional visualizations and quantitative results for the projector-based physical experiments discussed in~\cref{subsec:transferability}.
}

\xiao{\subsection{Cross-Location Transferability}
\label{subsec:supp_cross_location_transferability}

To support our claim in~\cref{sec:intro} that scene-level camouflage is location-invariant and can be applied consistently across an entire scene, we evaluate whether the learned camouflage generalizes to different locations within the same scene type. We conduct experiments on the COCO dataset targeting Faster R-CNN. Specifically, for each scene type, we select two vehicles and composite them into five additional backgrounds using PhotoShop by replacing only the background while keeping the vehicle appearance unchanged, ensuring that the resulting images remain visually natural. The new scenes introduce diverse environmental conditions, including different weather (\eg, cloudy, rainy, and foggy) and seasonal variations (\eg, summer and winter). Representative examples are shown in~\cref{fig:supp_transferable_other_places}. As illustrated, Faster R-CNN achieves an $\mathrm{AP}_{50}$ of 99.5\% on the corresponding clean images, whereas the camouflaged versions reduce performance to 38.2\%, corresponding to a 61.3\% drop. These results indicate that the learned scene-level camouflage remains adversarial even when the vehicle is placed at different locations within the same scene category, demonstrating strong cross-location transferability under diverse environmental conditions.
}

\begin{figure}[!tb]
    \centering  \includegraphics[width=0.8\columnwidth]{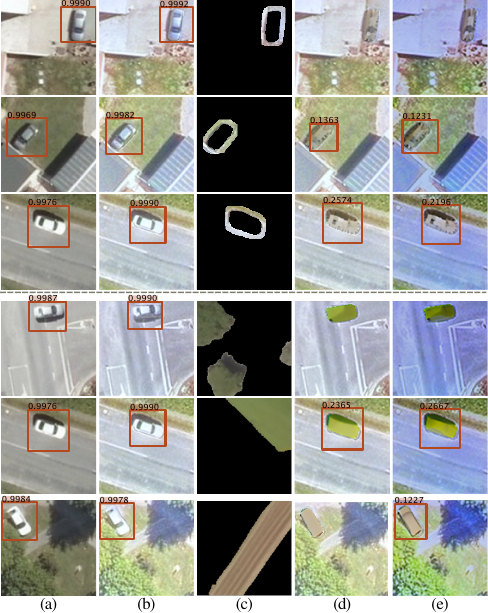}
    \caption{
    \textbf{Projector-based physical experiment for the LINZ dataset.} The first three rows present examples produced using the image-level strategy, and the last three rows illustrate the scene-level strategy. (a) Real images in digital space. (b) Photos captured from real images. (c) Reference areas used for style guidance.  (d) Camouflaged images generated from the captured photos.  (e) Photos taken after projecting the camouflaged images back onto the 3D car models.
    }
    \label{fig:supp_projector_linz}
\end{figure}

\begin{figure}[!tb]
    \centering  \includegraphics[width=0.8\columnwidth]{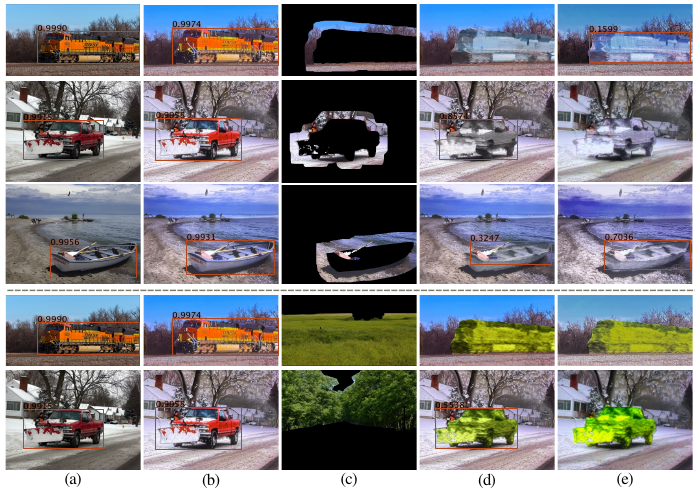}
    \caption{
    \textbf{Projector-based physical experiment for the COCO dataset.} The first three rows present examples produced using the image-level strategy, and the last three rows illustrate the scene-level strategy. (a) Real images in digital space. (b) Photos captured from real images projected on the 3D-printed scenes. (c) Reference areas used for style guidance.  (d) Camouflaged images generated from the captured photos.  (e) Photos taken after projecting the camouflaged images back onto the 3D-printed scenes.
    }
    \label{fig:supp_projector_coco}
\end{figure}

\subsection{Projector-based Physical Experiment}
\label{subsec:supp_physical_experiment}

\xiao{As discussed in \cref{subsec:transferability}, we conduct a projector-based physical test on both the LINZ and COCO dataset to examine the real-world transferability of the proposed camouflage. We provide the complete device configuration used in projector-based evaluation. As shown in~\cref{fig:supp_projector_linz_setup} (a) and~\cref{fig:supp_projector_coco_setup} (a), we employ an Epson EpiqVision\textsuperscript{TM} Mini EF12 projector to project real images and use an iPhone 16 Pro Max to capture the resulting scenes. The phone is positioned close to both the projector’s optical axis and the intended camera center to minimize parallax and perspective distortion, ensuring the captured photos faithfully reflect the viewpoint of a real detector. For the LINZ dataset, real-world images are projected onto a whiteboard, and a 3D-printed sedan model is placed at the corresponding vehicle location, as illustrated in~\cref{fig:supp_projector_linz_setup} (b). For the COCO dataset, images are projected directly onto 3D-printed scenes reconstructed from single-view images, as shown in~\cref{fig:supp_projector_coco_setup}. All physical models are printed using PLA matte filament, with a final length of 180 mm. All experiments are conducted under near-dark lighting to maximize projection contrast. Finally, ~\cref{fig:supp_projector_linz_setup}(c) shows an additional viewpoint illustrating how the 3D car model is affixed to the whiteboard, while ~\cref{fig:supp_projector_coco_setup} (c-d) provide additional views of the projection results on the reconstructed 3D scenes.

For the LINZ dataset, we evaluate five images for the scene-level strategy and four for the image-level strategy. For the COCO dataset, we evaluate three images for each strategy. Additional visualizations are shown in~\cref{fig:supp_projector_linz} and~\cref{fig:supp_projector_coco}. Because the number of physical test images is small, reporting $\mathrm{AP}_{50}$ is unreliable. With limited samples, a single prediction can significantly influence the precision–recall curve, leading to unstable estimates of detector performance. Therefore, we instead report the \textit{attack success rate}, which depends only on the detector confidence associated with the ground-truth object.

To determine a reliable confidence threshold, we evaluate the detector on the validation set and select the confidence value that yields the best $\mathrm{F1}$ score. The resulting threshold is 0.948 for LINZ and 0.806 for COCO. A detection is considered positive only if its confidence exceeds the threshold and its bounding box has IoU greater than 0.5 with the ground-truth vehicle.
 
Under this criterion, all LINZ samples produce confidence scores below 0.55 for scene-level, and below 0.22 for image-level camouflage. Both values are far below the 0.948 threshold.  For COCO samples, scene-level camouflaged vehicles are not detected, while image-level predictions have confidence scores below 0.71, also below the 0.806 threshold.
As a result, all physical test samples are suppressed by the detector, yielding an attack success rate of 100\% for both stylization strategies on both datasets. These results support the argument that camouflage patterns learned in simulation have the potential to transfer to real-world physical environments.
}

\begin{figure}[!tb]
    \centering  
    \includegraphics[width=0.98\columnwidth]{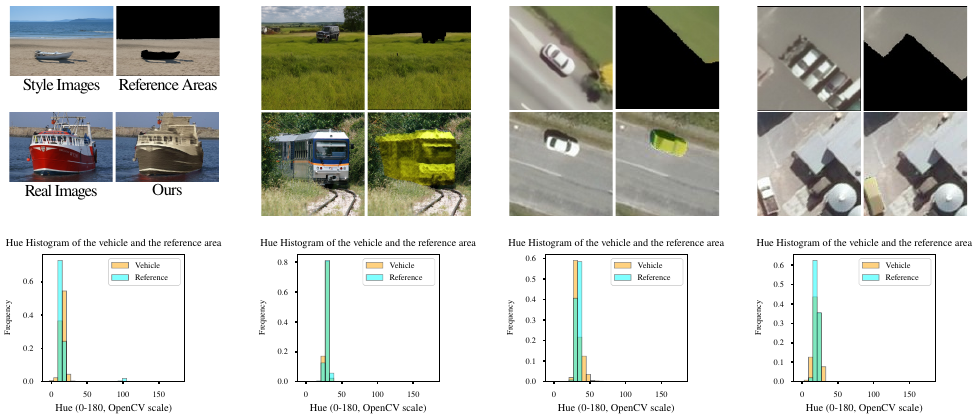}
    \caption{\textbf{Comparison of camouflaged vehicles and their corresponding style reference areas in HSV hue space.}
    }
    \label{fig:supp_failure_style}
\end{figure}

\begin{figure}[!tb]
    \centering  
    \includegraphics[width=0.65\columnwidth]{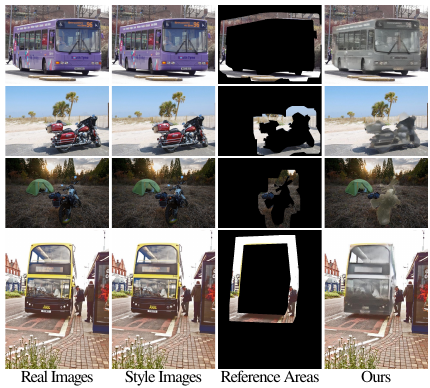}
    \caption{\textbf{Failure cases of the image-level camouflage generation strategy on the COCO dataset.}
    }
    \label{fig:supp_failure_coco}
\end{figure}

\section{Limitations}
\label{sec:supp_limitations}
First, our current pipeline relies on the L channel in LAB space as a coarse approximation for the shading component, which limits its ability to fully reproduce the true physical structure of the vehicle. Consequently, the camouflaged vehicles do not perfectly retain their original shading patterns, as observed in \cref{fig:supp_linz_scene_level_group2}, \cref{fig:supp_linz_scene_level}, and \cref{fig:supp_linz_image_level}. Since the L channel also carries texture information, the joint optimization of the structure preservation and style losses may interact during training, introducing subtle shifts in the vehicle’s final appearance. These effects are visible in~\cref{fig:supp_linz_scene_level_group2} and \cref{fig:supp_failure_style}, where the camouflaged vehicles exhibit perceptible differences relative to reference areas. Nonetheless, the accompanying hue histograms show strong alignment between the hue distributions of the camouflaged vehicles and their reference areas. Since hue specifies the angular position on the color wheel, representing the underlying color class independent of its value or saturation, this alignment indicates that our method reliably transfers the intended color component of the style, even when other perceptual attributes such as shading or saturation may be affected by the interactions of loss terms.

Additionally, the image-level strategy is less effective for ground-view images. Because vehicles in ground-view scenes exhibit strong perspective distortion, the dilated region surrounding the vehicle mask frequently extends into areas that are spatially distant or semantically unrelated to the target vehicle. As a result, the extracted reference regions provide inconsistent or misleading style guidance. Failure cases produced after the No-Box Attack stage are shown in \cref{fig:supp_failure_coco}, where the transferred vehicle's appearance does not align well with its immediate context. \xiao{These observations suggest that incorporating geometry-aware reference selection or perspective-aware style guidance could further improve performance in ground-view scenarios, which we leave for future work.}

\end{document}